\documentclass[10pt,journal,compsoc]{IEEEtran}
\ifCLASSOPTIONcompsoc
  \usepackage[nocompress,noadjust]{cite}
\else
  \usepackage[noadjust]{cite}
\fi
\ifCLASSINFOpdf
\else
\fi
\usepackage{amsmath,amssymb,amsfonts}
\usepackage{stfloats}
\usepackage{algpseudocode,algorithm,algorithmicx}
\algrenewcommand\algorithmicrequire{\textbf{Inputs:}}  
\algrenewcommand\algorithmicensure{\textbf{Outputs:}}
\renewcommand{\algorithmiccomment}[1]{\bgroup\hfill//~#1\egroup}
\usepackage{graphicx}
\usepackage{float}
\usepackage{textcomp}
\usepackage{tablefootnote}
\usepackage{hyperref}
\usepackage[perpage]{footmisc}
\usepackage{xcolor}
\usepackage{xspace}
\usepackage{booktabs}
\usepackage{paralist}
\usepackage{bbm}

\usepackage{dsfont}
\newcommand{\ind}{\mathds{1}}

\usepackage{subfig}

\newcommand{\method}{{\sc ECOD}\xspace}

\begin{document}
%
% paper title
% Titles are generally capitalized except for words such as a, an, and, as,
% at, but, by, for, in, nor, of, on, or, the, to and up, which are usually
% not capitalized unless they are the first or last word of the title.
% Linebreaks \\ can be used within to get better formatting as desired.
% Do not put math or special symbols in the title.
\title{\method: Unsupervised Outlier Detection Using Empirical Cumulative Distribution Functions}
%
%
% author names and IEEE memberships
% note positions of commas and nonbreaking spaces ( ~ ) LaTeX will not break
% a structure at a ~ so this keeps an author's name from being broken across
% two lines.
% use \thanks{} to gain access to the first footnote area
% a separate \thanks must be used for each paragraph as LaTeX2e's \thanks
% was not built to handle multiple paragraphs
%
%
%\IEEEcompsocitemizethanks is a special \thanks that produces the bulleted
% lists the Computer Society journals use for "first footnote" author
% affiliations. Use \IEEEcompsocthanksitem which works much like \item
% for each affiliation group. When not in compsoc mode,
% \IEEEcompsocitemizethanks becomes like \thanks and
% \IEEEcompsocthanksitem becomes a line break with idention. This
% facilitates dual compilation, although admittedly the differences in the
% desired content of \author between the different types of papers makes a
% one-size-fits-all approach a daunting prospect. For instance, compsoc 
% journal papers have the author affiliations above the "Manuscript
% received ..."  text while in non-compsoc journals this is reversed. Sigh.

% \textsuperscript{\textsection}
\author{Zheng~Li*,
        Yue~Zhao*\thanks{Y.~Zhao is the corresponding author.}, 
        ~\IEEEmembership{Student Member}
        Xiyang~Hu,
        Nicola~Botta,
        Cezar~Ionescu,
        and~George~H.~Chen% <-this % stops a space
\IEEEcompsocitemizethanks{
\IEEEcompsocthanksitem Z.~Li is with Northeastern University-Toronto and Arima Inc., Toronto, Canada, M5G. E-mail: winston@arimadata.com
% note need leading \protect in front of \\ to get a newline within \thanks as
% \\ is fragile and will error, could use \hfil\break instead.
\IEEEcompsocthanksitem Y.~Zhao, X.~Hu, and G.~H.~Chen are with the Heinz College of Information Systems and Public Policy, Carnegie Mellon University, Pittsburgh, PA, 15213. E-mail: \{zhaoy,xiyanghu,georgechen\}@cmu.edu
\IEEEcompsocthanksitem N.~Botta is with Potsdam Institute for Climate Impact Research, Potsdam, Germany, 14473. E-mail: botta@pik-potsdam.de
\IEEEcompsocthanksitem C.~Ionescu is with Deggendorf Institute of Technology, Deggendorf, Germany, 94469. E-mail: cezar.ionescu@th-deg.de
\IEEEcompsocthanksitem *Z.~Li and Y.~Zhao contributed equally with alphabetical ordering \protect\\}% <-this % stops an unwanted space
% \thanks{Manuscript received April 30, 2021; revised December 31, 2021;
% }
\thanks{Received April 30, 2021; revised December 31, 2021; accepted March 4, 2022}
}

\IEEEtitleabstractindextext{%

\begin{abstract}
Outlier detection refers to the identification of data points that deviate from a general data distribution. Existing unsupervised approaches often suffer from high  computational cost, complex hyperparameter tuning, and limited interpretability, especially when working with large, high-dimensional datasets. To address these issues, we present a simple yet effective algorithm called \method (Empirical-Cumulative-distribution-based Outlier Detection), which is inspired by the fact that outliers are often the ``rare events" that appear in the 
tails of a distribution. 
In a nutshell, \method first estimates the 
underlying distribution of the input data in a nonparametric fashion by computing the empirical cumulative distribution per dimension of the data. \method then uses these empirical distributions to estimate tail probabilities per dimension for each data point. Finally, \method computes an outlier score of each data point by aggregating estimated tail probabilities across dimensions. 
Our contributions are as follows:
(1) we propose a novel outlier detection method called \method, which is both parameter-free and easy to interpret;
(2) we perform extensive experiments on 30 benchmark datasets, where we find that \method outperforms 11 state-of-the-art baselines in terms of accuracy, efficiency, and scalability; 
and (3) we release an easy-to-use and scalable (with distributed support) Python implementation for accessibility and reproducibility.
\end{abstract}

% Note that keywords are not normally used for peerreview papers.
}

% make the title area
\maketitle
% \begingroup\renewcommand\thefootnote{\textsection}
% \footnotetext{Equal contribution}
% \endgroup

% To allow for easy dual compilation without having to reenter the
% abstract/keywords data, the \IEEEtitleabstractindextext text will
% not be used in maketitle, but will appear (i.e., to be "transported")
% here as \IEEEdisplaynontitleabstractindextext when the compsoc 
% or transmag modes are not selected <OR> if conference mode is selected 
% - because all conference papers position the abstract like regular
% papers do.
\IEEEdisplaynontitleabstractindextext
% \IEEEdisplaynontitleabstractindextext has no effect when using
% compsoc or transmag under a non-conference mode.

% For peer review papers, you can put extra information on the cover
% page as needed:
% \ifCLASSOPTIONpeerreview
% \begin{center} \bfseries EDICS Category: 3-BBND \end{center}
% \fi
%
% For peerreview papers, this IEEEtran command inserts a page break and
% creates the second title. It will be ignored for other modes.
\IEEEpeerreviewmaketitle

\IEEEraisesectionheading{
\section{Introduction}
\label{sec:introduction}}

\IEEEPARstart{O}{utliers}, also sometimes referred to as anomalies, are data points with different data characteristics from ``normal'' observations. Since outliers can markedly impact the results of statistical analyses, removing outliers is often a crucial preprocessing step in data analysis models (\cite{yang2014robust,najafi2019outlier}). However, to remove outliers, we need to identify them first, which is the goal of outlier detection (OD). OD has many applications, such as fraud detection (\cite{cao2018collective,tao2019mvan,dou2020enhancing}), network intrusion detection (\cite{Chandola2009,zhang2015novel}), social media analysis (\cite{yu2017ring,zhao2020multi}), video analysis \cite{cui2007sequential}, intelligent transportation \cite{chen2010comparison,xu2021opencda,xu2021opv2v}, and data generation \cite{wan2020sync}. These applications often demand
OD algorithms with high detection accuracy and fast execution while being easy to interpret.

Numerous unsupervised OD algorithms have been proposed over the years (e.g., \cite{Ramaswamy2000efficient,scholkopf2001estimating,liu2008isolation,zhao2019lscp,DBLP:conf/icml/RuffGDSVBMK18,Liu2019generative}; we provide a more detailed overview in Section~\ref{sec:related_work}).
These existing approaches have a few limitations. First of all, many of these methods, especially the ones requiring density estimation and pairwise distance calculation, suffer from the curse of dimensionality---both detection accuracy and runtime efficiency worsen rapidly as the number of data points and their dimensionality increase (\cite{zhao2021suod,zhao2021tod}).
Second, most methods require hyperparameter tuning, 
which is difficult in the unsupervised setting (\cite{zhao2021automatic,DBLP:conf/ijcai/Akoglu21}).

To address these limitations, we propose a simple yet effective method using \textbf{e}mpirical \textbf{c}umulative distribution functions for \textbf{o}utlier \textbf{d}etection, abbreviated as \method. 
\method is motivated by the definition of outliers, which may be viewed as the \textit{rare events} in the data (\cite{Lazarevic2005feature,pokrajac2007incremental}). Rare events are often the ones that appear in one of the tails of a distribution. As a concrete example, if the data are sampled from a one-dimensional Gaussian, then points in the left or right tails are often considered rare events or outliers. This has motivated commonly used heuristic OD approaches such as the ``three-sigma'' rule that declares points more than three standard deviations from the mean to be outliers  (\cite{leys2013detecting,bakar2006comparative}; there's also a robust variant called the ``1.5 IQR'' rule \cite{dekking2005modern}). However, the three-sigma rule only uses the mean and standard deviation of a distribution. Instead, one could capture more information of the distribution by building a histogram of the data and using bins with low counts to determine where outliers are \cite{goldstein2012histogram}. However, such an approach requires tuning over different ways to bin the data for histogram construction. Our approach \method avoids this problem of tuning altogether by estimating the empirical cumulative distribution function (ECDF) of the data, which has no parameters to tune and approximates the entire distribution without making any parametric assumptions.

The technical difficulty in using ECDFs arises when working with high-dimensional data: the joint ECDF over all variables converges more slowly to the true joint CDF as the number of dimensions increases \cite{naaman2021tight}. Our approach \method sidesteps this issue in a straightforward manner: we compute a univariate ECDF for each dimension separately. Then to measure the outlyingness of a data point, we compute its tail probability across all dimensions via an independence assumption, which amounts to multiplying all the estimated tail probabilities from the different univariate ECDFs. We do this calculation in log space and in a manner that accounts for both the left and right tails of each of the dimensions. Despite this independence assumption appearing to be quite strong, \method turns out to work very well in practice.

In summary, we propose a novel OD approach \method that has the following key advantages:
\begin{itemize}
    \item \textbf{Effectiveness}: Through extensive evaluation, we show that \method outperforms 11 popular baseline OD methods on 30 benchmarks. Specifically, \method ranks the highest, and scores 2\% higher in the area under the receiver operating characteristic curve and 5\% higher in average precision than the second best detector.
    \item \textbf{Efficiency and scalability}: \method has time complexity $\mathcal{O}(nd)$, where $n$ is the number of data points and $d$ is the number of dimensions, and can trivially be parallelized across dimensions. Moreover, since \method has no hyperparameters, there is no time spent on hyperparameter tuning. With a single thread, \method can handle datasets with 1,000,000 observations and 10,000 features on a standard personal laptop in 2 hours.
    \item \textbf{Interpretability}: \method is easy to interpret. For any data point, we can look at its left or right estimated tail probability per dimension. This tells us how each dimension contributes to the overall outlier score we use. 
    This information guides practitioners regarding which dimensions to focus on for improving data quality.
\end{itemize}
The rest of this paper is organized as follows:  we review some existing OD techniques and their strengths and limitations in Section \ref{sec:related_work}.
We describe the algorithmic design of \method and its properties in  Section \ref{sec:proposed_algorithm}. We carry out experiments to compare \method with state-of-the-art OD methods and demonstrate that it is one of the most accurate, efficient, and scalable methods in Section \ref{sec:empirical_evaluation}. 
Finally, we conclude the paper in Section \ref{sec:conclusion} with future directions, including a discussion on how to remove the independence assumption in how we aggregate tail probability information across dimensions. To facilitate reproducibility and accessibility, we open-source \method with distributed learning support as part of the popular PyOD library\footnote{Implementation in PyOD library: \url{https://github.com/yzhao062/pyod/blob/master/pyod/models/ecod.py}}.

\section{Related Work}
\label{sec:related_work}
Over the years, different types of unsupervised outlier detection algorithms have been proposed (\cite{aggarwal2015outlier,pang2021deep}). In this section, we give an overview of key methods for tabular data. Among them, eleven state-of-the-art algorithms are selected as baselines for our experiments in Section \ref{sec:empirical_evaluation}.

\subsection{Proximity-based Algorithms}
First, proximity-based algorithms, as their name suggests, are based on local neighborhood information around each point. These algorithms can be categorized into density- and distance-based algorithms \cite{aggarwal2015outlier}.

The core principle behind density-based outlier detection methods is that an outlier can be found in a low-density region, whereas non-outliers (also called \emph{inliers}) are assumed to appear in dense neighborhoods. For example, the local outlier factor, proposed by Breunig \textit{et al.} \cite{Breunig2000lof}, is based on the concept of local density, where locality is given by the $k$ nearest neighbors. Distances of these nearest neighbors are then used to help estimate local densities. We can declare a data point with substantially lower local density than its neighbors to be an outlier. A few improvements were later introduced, including the Connective-based Outlier Factor (COF) by Tang \textit{et al.}~\cite{Tang2002cof}, and LOcal Correlation Integral (LOCI) by Papadimitriou \textit{et al.}~\cite{Papadimitriou2003loci}.

Distance-based methods compare objects to their neighbors, and those considerably far from their neighbors are deemed outliers. This procedure does not estimate local densities. In determining which points are considered neighbors, commonly the $k$ nearest neighbors are used \cite{Ramaswamy2000efficient}.

Proximity-based methods are mostly nonparametric as they do not assume any parametric distribution for the data. This can be a key advantage in outlier detection as minimal prior knowledge is required on the probability distribution of the dataset. This makes them intuitive and easy to understand. However, this family of methods is typically computationally expensive, sensitive to hyperparameters such as how to define the neighbors (e.g., which distance/similarity function to use, how many neighbors to consider), and vulnerable to the curse of dimensionality. Our proposed algorithm \method address both drawbacks---it has no hyperparameters to tune, and has time complexity that scales linearly in the dataset size and dimensions.

\subsection{Statistical Models}
In the context of OD, approaches based on statistical models first fit probability distributions to data points. They then determine whether points are outliers based on the fitted models. These approaches are usually categorized into two main groups---parametric and nonparametric methods. The key difference is that parametric methods assume that the data come from a parametric distribution, so fitting such a distribution amounts to learning the parameters of the assumed parametric distribution. Common parametric methods for OD include using Gaussian mixture models (GMM) \cite{yang2009gmm} and linear regression \cite{satman2013reg}. In contrast, nonparametric methods do not assume a parametric model for the data. Some examples include Kernel Density Estimation (KDE) \cite{pavlidou2014kde}, histogram-based methods (HBOS) \cite{goldstein2012histogram}, and a few other variants. Typically, parametric models for OD are fast to use after the model fitting step, whereas nonparametric models for OD can be more expensive to work (for example, depending on the kernel used for KDE, the fitted model could---for an exact solution---require comparing a candidate test data point with all training data in deciding whether the test point is an outlier or not).

Note that our proposed algorithm \method uses a statistical model that is nonparametric although this model cannot represent all possible multivariate distributions. In particular, we model each dimension of the data in a fully nonparametric fashion (with a univariate ECDF) but aggregate information across dimensions. We assume that the different dimensions are independent. Thus, even though our approach has no parameters, it does make a strong structural assumption on the underlying joint probability distribution over the different dimensions/features.

\subsection{Learning-based Models}
Learning-based approaches involve training machine learning models to predict which points are outliers. Based on a training set, a classification model is trained to classify outliers and inliers. The same model predicts which points in the test set are outliers. A classical method in this category is the one-class SVM (OCSVM) \cite{scholkopf2001estimating}. Another popular method leverages clustering to model the behavior of data points \cite{he2003discovering}, and identifies outliers based on cluster assignments. 

Using neural networks for detecting outliers has gained more attention in the last decade. Some representative examples from this line of work include the use of generative adversarial networks (GANs) \cite{Liu2019generative}, autoencoders \cite{chen2017outlier} (including variational autoencoders), and reinforcement learning \cite{pang2020toward}. Recent developments in this direction can be found in the survey by Pang \textit{et al.}~\cite{pang2021deep}.

Learning-based methods work well in practice on large datasets.
However, most learning-based methods are computationally expensive.
Moreover, they tend to involve nontrivial hyperparameter tuning, especially in the unsupervised setting. Lastly, state-of-the-art learning-based methods that use deep nets are often difficult to interpret.

\subsection{Ensemble-based models}
Ensemble-based approaches in OD combine results from various base outlier detectors to produce more robust OD results. The intuition is similar to how ensembling works in standard classification/regression tasks (e.g., bagging, boosting, random forests). Notable works of ensemble-based OD include feature bagging \cite{Lazarevic2005feature} that uses various sub-feature spaces, isolation forests \cite{liu2008isolation} that aggregate the information from multiple base trees, LSCP \cite{zhao2019lscp} that dynamically picks the best base estimator for each data point, and SUOD \cite{zhao2021suod} that uses many heterogeneous estimators.

In general, ensemble-based methods for OD often work well in practice even for high-dimensional datasets. However, these methods also can involve nontrivial tuning, such as in selecting the right meta-detectors \cite{zhao2021automatic}. Additionally, ensemble-based methods are often less interpretable.

Our proposed algorithm \method could be considered an ensemble model in that we are learning a nonparametric statistical model per dimension of the data, and then we aggregate/ensemble the models across dimensions to detect outliers. Because each ``base'' model uses only a single dimension of the data and we straightforwardly aggregate the base models, \method is easy to interpret.

\section{Proposed Algorithm: \method}
\label{sec:proposed_algorithm}
We now present the details of \method in three subsections. First, we provide a problem statement of unsupervised outlier detection and the associated challenges in Section \ref{subsec:problem_statement}. We then provide the motivation and technical details of \method in Section \ref{subsec:ecod}. We discuss properties of \method in Section \ref{subsec:ecod_property}, including interpretability and scalability.

\subsection{Problem Formulation and Challenges}
\label{subsec:problem_statement}

We consider the following standard unsupervised outlier detection setup. We assume that we have $n$ data points $X_1,X_2,\dots,X_n\in\mathbb{R}^d$ that are sampled i.i.d. We collectively refer to this entire dataset by the matrix $\mathbf{X}\in\mathbb{R}^{n\times d}$, which is formed by stacking the different data points' vectors as rows. Given $\mathbf{X}$, an OD model $M$ assigns, for each data point $X_i$, an outlier score $O_i\in\mathbb{R}$ (higher means more likely an outlier).

There are a few notable challenges for unsupervised outlier detection:
\begin{itemize}
    \item \textit{Curse of dimensionality}: many OD algorithms described in Section \ref{sec:related_work} become less accurate or have computational scalability issues when the input dataset $\mathbf{X}$ is high-dimensional ($d$ is either larger than or of a similar order of magnitude as $n$), or the number of data points $n$ is large. Specifically for proximity-based methods, density estimation and (pairwise) distance calculation become more computationally expensive \cite{zhao2021suod}. Moreover, in general, density estimation in the high-dimensional setting can require the number of data points to scale exponentially in the number of dimensions \cite{stone1980optimal}.
    \item \textit{Limited interpretability}: OD applications often require interpretability/explainability to a certain extent. For instance, in detecting whether a financial transaction is an outlier and considered ``fraudulent'', it could be helpful to provide some sort of evidence as to why a transaction is considered fraudulent or not. This capacity is often absent in most existing OD algorithms we are aware of.
    \item \textit{Complexity in hyperparameter tuning}: without access to any ground truth labels for which data points are outliers, model selection and hyperparameter tuning are challenging in existing OD algorithms \cite{zhao2021automatic}.
\end{itemize}

\begin{algorithm} [ht]
	\caption{Unsupervised OD Using ECDF (\method)}
	\label{alg:ECOD}
	\begin{algorithmic}[1]
 		\Require{input data $\mathbf{X} =\{X_i\}_{i=1}^n
 		\in \mathbbm{R}^{n \times d}$ with $n$ samples and $d$ features; $X_i^{(j)}$ refers to the value of $j$-th feature of the $i$-th sample
		} 
		\Ensure{outlier scores $\mathbf{O} := \text{ECOD}(\mathbf{X}) \in \mathbbm{R}^{n}$}
		\For{each dimension $j$ in $1,\ldots,d$}
		\State Estimate left and right tail ECDFs (using equations~\eqref{eq:univariate-ECDF-left} and~\ref{eq:univariate-ECDF-right}, which we reproduce below):
		\begin{align}
		    \text{left tail ECDF: }
		    \widehat{F}_{\textsf{left}}^{(j)}(z)
= \frac{1}{n}\sum_{i=1}^n \ind\{X_i^{(j)}\le z\}\text{ for }z\in\mathbb{R}, \nonumber\\
		    \text{right tail ECDF: }
		    \widehat{F}_{\textsf{right}}^{(j)}(z)
= \frac{1}{n}\sum_{i=1}^n \ind\{X_i^{(j)}\ge z\}\text{ for }z\in\mathbb{R}. \nonumber
		\end{align}
		\State Compute the sample skewness coefficient for the \mbox{$j$-th} feature's distribution:
		$$\gamma_j = \frac{\tfrac{1}{n} \sum_{i=1}^n (X_i^{(j)}-\overline{X^{(j)}})^3}{ \big[\tfrac{1}{n-1} \sum_{i=1}^n (X_i^{(j)}-\overline{X^{(j)}})^2\big]^{3/2}},$$
		where $\overline{X^{(j)}}=\frac{1}{n}\sum_{i=1}^n X_i^{(j)}$ is the sample mean of the $j$-th feature.
		\EndFor
		
		\For{each sample $i$ in $1,\ldots, n$}
	    \State Aggregate tail probabilities of $X_i$ to obtain outlier score $O_i$: 
	    \Comment{\S \ref{subsubsec:outlier_scores}}
		\begin{align}
		    O_{\textsf{left-only}}(X_i) &= -\sum_{j=1}^d \log( \widehat{F}_{\textsf{left}}^{(j)}(X_i^{(j)}) ), \nonumber\\
		    O_{\textsf{right-only}}(X_i) &= -\sum_{j=1}^d \log( \widehat{F}_{\textsf{right}}^{(j)}(X_i^{(j)}) ), \nonumber\\
		    O_{\textsf{auto}}(X_i) &=
		    -\sum_{j=1}^d
		    [
		    \ind\{\gamma_j < 0\}
		    \log( \widehat{F}_{\textsf{left}}^{(j)}(X_i^{(j)}) ) \nonumber \\
		    & \phantom{=-\sum_{j=1}^d[}
		    +
		    \ind\{\gamma_j \ge 0\}
		    \log( \widehat{F}_{\textsf{right}}^{(j)}(X_i^{(j)}) ) \nonumber
		    ].
		\end{align}
	    \State Set the final outlier score for point $X_i$ to be $$O_i = \max\{O_{\textsf{left-only}}(X_i), O_{\textsf{right-only}}(X_i), O_{\textsf{auto}}(X_i)\}.$$
		\EndFor
		\State \textbf{Return} outlier scores $\mathbf{O}=(O_1,\dots,O_n)$
	\end{algorithmic}
\end{algorithm}

\subsection{The Proposed \method }
\label{subsec:ecod}

\subsubsection{Motivation and High-level Idea}
\label{subsubsec:rare_events}

A natural way to characterize outliers is to take them to correspond to rare events that occur in low-density parts of probability distribution (\cite{Lazarevic2005feature,pokrajac2007incremental}). If the distribution is unimodal, then these rare events occur in the tails of the distribution. With this motivation, for each observation $X_i$, our method \method is based on computing the probability of observing a point at least as ``extreme'' as $X_i$ in terms of tail probabilities.

Specifically, let $F:\mathbb{R}^d\rightarrow[0,1]$ denote the joint cumulative distribution function (CDF) across all $d$ dimensions/features. In particular, $X_1,X_2,\dots,X_n$ are assumed to be sampled i.i.d.~from a distribution with joint CDF $F$. For a vector $z\in\mathbb{R}^d$, we denote its $j$-th entry as $z^{(j)}$, e.g., we write the $j$-th entry of $X_i$ as $X_i^{(j)}$. We use the random variable $X$ to denote a generic random variable with the same distribution as each $X_i$. Then by the definition of a joint CDF, for any $x\in\mathbb{R}^d$,
\[
F(x) = \mathbb{P}(\underbrace{X^{(1)} \le x^{(1)}, X^{(2)} \le x^{(2)}, \dots, X^{(d)} \le x^{(d)}}_{\text{abbreviated as ``}X\le x\text{''}})
\]
This probability is a measure of how ``extreme'' $X_i$ is in terms of left tails: the smaller $F(X_i)$ is, then the less likely a point $X$ sampled from the same distribution as $X_i$ will satisfy the inequality $X \le X_i$ (again, this inequality needs to hold across all $d$ dimensions). Similarly, $1-F(X_i)$ is also a measure of how ``extreme'' $X_i$ is, however, looking at the right tails of every dimension instead of the left tails. Therefore, if either $F_\mathbf{X}(X_i)$ or $1 - F_\mathbf{X}(X_i)$ is extremely small, then this suggests that $X_i$ corresponds to a rare event and is, therefore, likely to be an outlier.

The challenge is that in practice, we do not know the true joint CDF and have to estimate it from data. The rate of convergence in estimating a joint CDF using a joint ECDF slows down as the number of dimensions increases \cite{naaman2021tight}. As a simplifying assumption, we assume that the different dimensions/features are independent so that joint CDF has the factorization
\[
F(x) = \prod_{j=1}^d F^{(j)}(x^{(j)})
\qquad\text{for }x\in\mathbb{R}^d,
\]
where $F^{(j)}:\mathbb{R}\rightarrow[0,1]$ denotes the univariate CDF of the $j$-th dimension: $F^{(j)}(z)=\mathbb{P}(X^{(j)}\le z)$ for $z\in\mathbb{R}$.

Now it suffices to note that univariate CDF's can be accurately estimated simply by using the empirical CDF (ECDF), namely:
\begin{equation}
\widehat{F}_{\textsf{left}}^{(j)}(z)
:= \frac{1}{n}\sum_{i=1}^n \ind\{X_i^{(j)}\le z\}
\qquad\text{for }z\in\mathbb{R},
\label{eq:univariate-ECDF-left}
\end{equation}
where $\ind\{\cdot\}$ is the indicator function that is~1 when its argument is true and is~0 otherwise.

Previously, we mentioned that the right tail could be obtained by looking at 1 minus a CDF. However, we remark that there is a slight asymmetry in doing this since
\[
1-F^{(j)}(z)=1-\mathbb{P}(X^{(j)}\le z)=\mathbb{P}(\underbrace{X^{(j)}> z}_{\text{note the strict inequality}}).
\]
Basically $F^{(j)}(z)$ does not use a strict inequality whereas $1-F^{(j)}(z)$ does. In how we aggregate tail probabilities, we will combine information from both left and right tails, and for symmetry, we actually separately also compute the ``right-tail'' ECDF:
\begin{equation}
\widehat{F}_{\textsf{right}}^{(j)}(z)
:= \frac{1}{n}\sum_{i=1}^n \ind\{X_i^{(j)}\ge z\}
\qquad\text{for }z\in\mathbb{R}.
\label{eq:univariate-ECDF-right}
\end{equation}
Thus, we can estimate the joint left and right-tail ECDFs across all $d$ dimensions under an independence assumption via the estimates
\begin{gather}
\widehat{F}_{\textsf{left}}(x)
=
\prod_{j=1}^d
\widehat{F}_{\textsf{left}}^{(j)}(x^{(j)})
\;\text{and}\;
\widehat{F}_{\textsf{right}}(x)
=
\prod_{j=1}^d
\widehat{F}_{\textsf{right}}^{(j)}(x^{(j)})
\nonumber \\
\text{for }x\in\mathbb{R}^d.
\label{eq:joint-ECDFs-indep}
\end{gather}
Our proposed method \method builds on the ideas we just presented and consists of two main steps: first, we compute each dimension's left- and right-tail ECDFs as given in equations~\eqref{eq:univariate-ECDF-left} and~\eqref{eq:univariate-ECDF-right}. Next, for every point $X_i$, we aggregate its tail probabilities $\widehat{F}_{\textsf{left}}^{(j)}(X_i^{(j)})$ and $\widehat{F}_{\textsf{right}}^{(j)}(X_i^{(j)})$ to come up with a final outlier score $O_i\in[0,\infty)$; higher means more likely to be an outlier. Note that these outlier scores are not probabilities and are meant to be used to compare the data points. We explain in more detail how we aggregate tail probabilities in Section~\ref{subsubsec:outlier_scores}. An important idea we use in aggregating tail probabilities is that it does not always make sense to only consider the left tail probability for every dimension and then separately only consider the right tail probability for every dimension. Meanwhile, especially when $d$ is large, it is impractical to consider all $2^d$ possible combinations of whether to use the left or right tail probability per dimension. Our aggregation step uses the skewness of a dimension's distribution to automatically select whether we use the left or the right tail probability for a dimension. The pseudocode of \method is given in Algorithm~\ref{alg:ECOD}.

\begin{figure*}[!htp] % [!htb]
\centering
\subfloat[Ground truth]{%
  \includegraphics[clip,width=0.41\columnwidth]{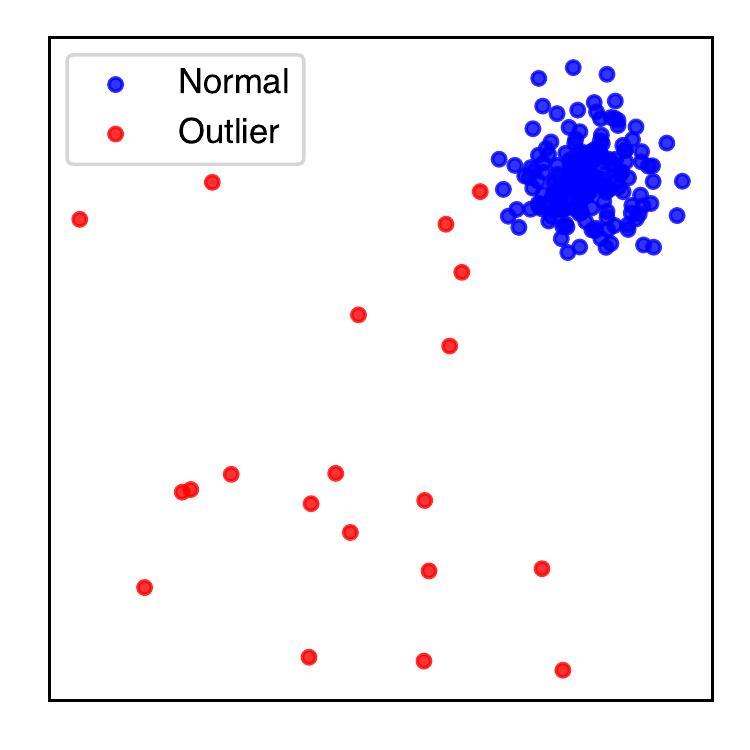}%
}
\subfloat[Left tail]{%
  \includegraphics[clip,width=0.41\columnwidth]{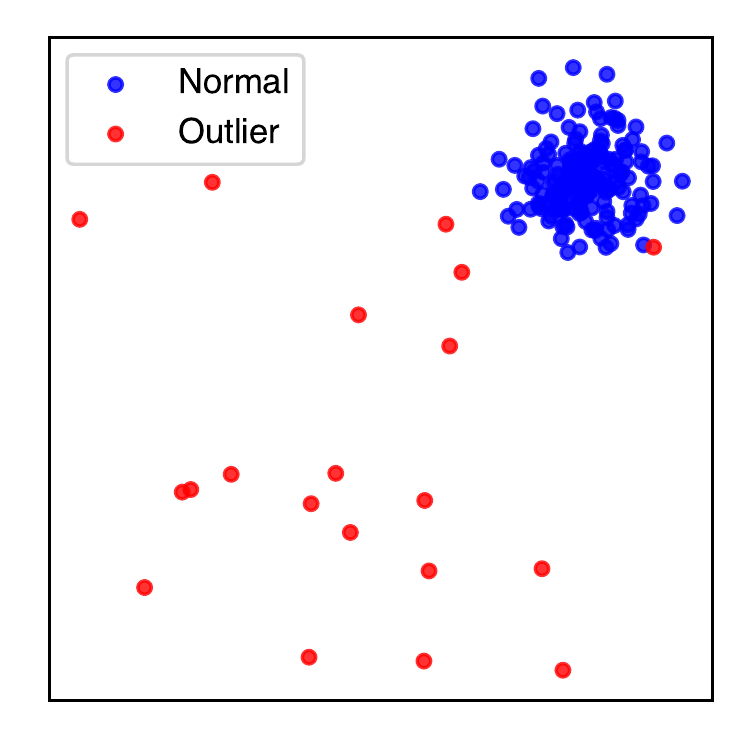}%
}
\subfloat[Right tail]{%
  \includegraphics[clip,width=0.41\columnwidth]{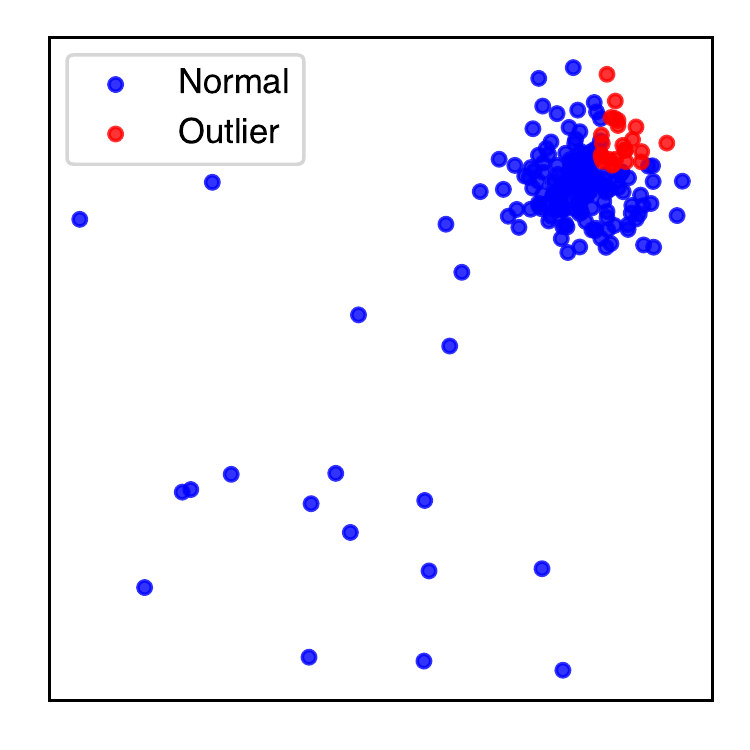}%
}
\subfloat[Avg of two tails]{%
  \includegraphics[clip,width=0.41\columnwidth]{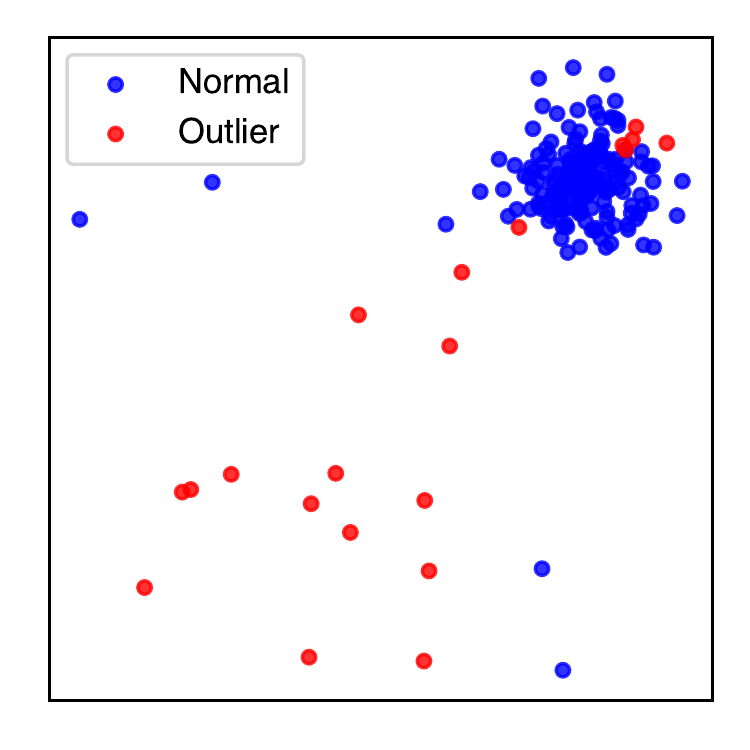}%
}
\subfloat[Skewness corrected]{%
  \includegraphics[clip,width=0.41\columnwidth]{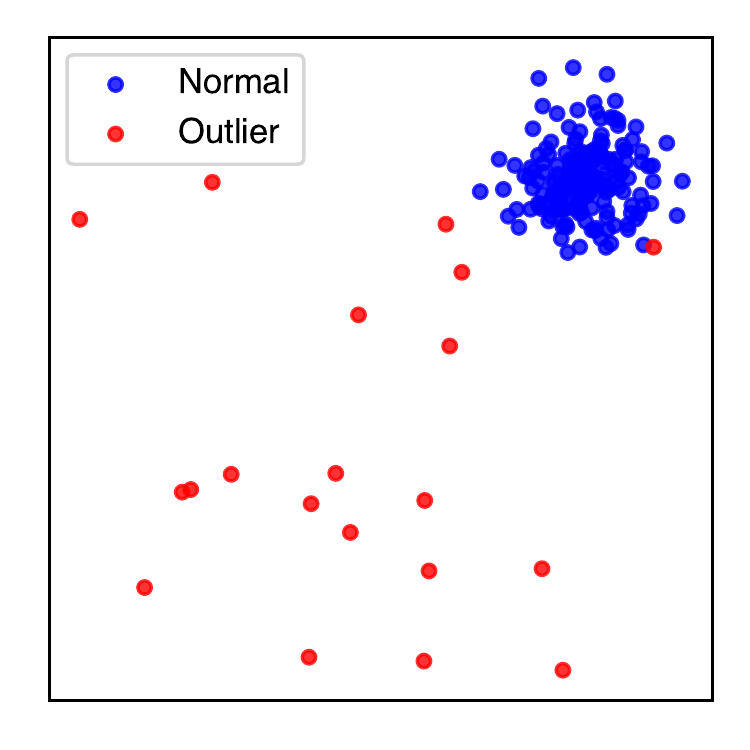}%
}
\\
\caption{An illustration of how using different tail probabilities affect the results. The leftmost column is plots of the ground truth; the second column is the result if left tail probabilities are used. The middle column corresponds to outlier detection if right tail probabilities are used, followed by the average of both tail probabilities in the fourth column and skewness corrected tail probabilities (SC) in the rightmost column (which shows the best result).}
\label{fig:tail_variant}
\end{figure*}

\subsubsection{Aggregating Outlier Scores} 
\label{subsubsec:outlier_scores}

\textbf{Using skewness to decide on whether to use the left or the right tail probability for a specific dimension.}
Using only the left tail probability for every dimension or only the right tail probability for every dimension can be limiting in finding outliers it could be that in a specific dimension, being in the left tail could be considered more of an outlier whereas in another dimension, we should instead consider the right tail. To automatically decide which tail we should consider for a particular dimension, we consider the skewness of that dimension's distribution.

Consider Fig.~\ref{fig:tail_variant}, where we demonstrate the effect of using different tail probabilities in catching outliers. The subfigure on the left-most column is the ground truth of the synthetic dataset, where outliers are labeled in red and the normal points are labeled in blue. The dataset is generated by taking 180 samples from a normal distribution centered at the top right corner of a unit square as inliers, combined with 20 samples from a uniform distribution on the square as outliers. In the next three columns, we show the results of using left tail probabilities, right tail probabilities, and averages of both tail probabilities to denote outlyingness, respectively. For each situation, we consider the same number of outliers (i.e., 20).

Because of the nature of this dataset (i.e., all outliers fall on the left end of the 2-dimensional distribution), using the left tail probabilities works surprising well (Fig.~\ref{fig:tail_variant}, column 2), because all outliers, by the construction of the dataset, are ``smaller" than inliers. On the other hand, using the right tail worked extremely poorly (Fig.~\ref{fig:tail_variant}, column 3), because there are no extremely large outliers. Thus, it would have wrongfully captured the relatively large points as outliers. Taking an average of the two tail probabilities (Fig.~\ref{fig:tail_variant}, column 4) compromises the situation, and identifies half of the left tail outliers and misidentifies half of the right tail. We should point out that if the dataset is flipped, and all outliers appear on the top right corner while the inliers appear on the bottom left, then using the right tail probabilities for identifying outliers would have worked perfectly, while left tail probabilities would perform poorly. Averaging, on the other hand, works if both large and small outliers occur.

To overcome this, we note that the skewness of the dataset plays a major role in whether left tail probabilities or right tail probabilities should be used. In this example, both marginals of $X_1$ and $X_2$ skew negatively (i.e., the left tail is longer and the mass of the distribution is concentrated on the right). In this case, it makes sense to use the left tail probabilities. For dimensions that skew positively, using the right tail probabilities would become a better choice.
The right-most column of Fig.~\ref{fig:tail_variant} demonstrates the outcome of \method of tail selection is based on the skewness of each dimension, and we can see that it is able to capture the correct outliers. We called this the skewness corrected (SC) version of \method and we compare  the performance of different variants of \method using different tail probabilities in Section \ref{subsec:variant_comparison}.

We, therefore, propose to decide which tail to use in \method based on the skewness of the underlying distribution, where the sample skewness coefficient of dimension $j$ can be calculated as below \cite{groeneveld1984measuring}:
$$\gamma_j = \frac{\tfrac{1}{n} \sum_{i=1}^n (X_i^{(j)}-\overline{X^{(j)}})^3}{ \big[\tfrac{1}{n-1} \sum_{i=1}^n (X_i^{(j)}-\overline{X^{(j)}})^2\big]^{3/2}},$$
where $\overline{X^{(j)}}=\frac{1}{n}\sum_{i=1}^n X_i^{(j)}$.
When $\gamma_j < 0$, we can thus consider points in the left tail to be more outlying. When $\gamma > 0$, we instead consider points in the right tail to be more outlying.

\smallskip
\noindent
\textbf{Final aggregation step.}
To compute the final outlier score per data point, we simply work in the negative log probability space. In particular, we compute equation~\eqref{eq:joint-ECDFs-indep} as:
\begin{align}
O_{\textsf{left-only}}(X_i)
&:=
-\log\widehat{F}_{\textsf{left}}(X_i)
= -\sum_{j=1}^d \log(\widehat{F}_{\textsf{left}}^{(j)}(X_i^{(j)})), \\ 
% \label{eq:left_only}
O_{\textsf{right-only}}(X_i)
&:=
-\log\widehat{F}_{\textsf{right}}(X_i)
= -\sum_{j=1}^d \log(\widehat{F}_{\textsf{right}}^{(j)}(X_i^{(j)})). 
\label{eq:right_only}
\end{align}
Meanwhile, we also compute the ``automatic'' version that decides whether to use the left or the right tail of the $j$-th dimension based on whether $\gamma_j < 0$ or $\gamma_j > 0$. Note that if the data are sampled from a continuous random variable, then $\gamma=0$ with probability 0, so for simplicity, we just break the tie in favor of one of the two directions. We specifically use the automatic outlier score:
\begin{align*}
O_{\textsf{auto}}(X_i) &=
-\sum_{j=1}^d
[
\ind\{\gamma_j < 0\}
\log( \widehat{F}_{\textsf{left}}^{(j)}(X_i^{(j)}) ) \nonumber \\
& \phantom{=-\sum_{j=1}^d[}
+
\ind\{\gamma_j \ge 0\}
\log( \widehat{F}_{\textsf{right}}^{(j)}(X_i^{(j)}) ).
\end{align*}
Since we are operating in negative log probability space, lower probability (which suggests a more rare occurrence) corresponds to higher negative log probability. We use whichever negative log probability score is highest as the final outlier score $O_i$ for point $X_i$, i.e.,
\begin{equation}
O_i = \max\{O_{\textsf{left-only}}(X_i), O_{\textsf{right-only}}(X_i), O_{\textsf{auto}}(X_i)\}.
\label{eq:outlier_scores}
\end{equation}
Essentially we are using the most extreme of the three scores computed.

\subsection{Properties of \method}
\label{subsec:ecod_property}
\subsubsection{\method as an Interpretable Outlier Detector}
\label{subsec:outlier_plot}
Interpretability in machine learning is an important concept, as it provides domain experts some insights into how algorithms make their decisions \cite{hu2019optimal,hu2021uncovering}. Interpretable algorithms provide both transparency and reliability. Having a transparent model means that humans can learn from the thought process of models, and try to discover the ``whys" behind why a particular data point is classified. 
Clearly, interpretability is also critical in outlier detection applications \cite{gupta2018beyond}. For instance, explaining why a transaction is fraudulent is equally important to identifying it.

As \method evaluates the outlying behaviors on a dimensional basis, we could use \method as an explainable detector for dimensional contribution. From equation \ref{eq:outlier_scores}, we know that if a sample has a large outlier score, then at least one of the three tail probabilities is large. Thus, let $O_i^{(j)}$
be the dimensional outlier score for dimension $j$ of $X_i$, and using the fact that the $\log$ function is monotonic, we can see that it represents the degree of outlyingness of dimension $j$. This can be compared against some preset thresholds, such as $-\log(0.01) = 4.61$, or top $\alpha$ percent of $O(X_i^{(j)}) \; \forall i$ to give practitioners some indications why particular points are considered outliers. By plotting its corresponding graph (i.e., \textit{Dimensional Outlier Graph}), a more direct understanding of features' contribution can be acquired.

\noindent \textbf{Illustration with Breast Cancer Wisconsin Dataset}. We demonstrate the interpretability aspect of \method with the Breast Cancer Wisconsin (Diagnostic) Data Set (\textit{Breastw}, Table \ref{table:data}) as an example. Cell samples are provided by Dr. William H. Wolberg from the University of Wisconsin as part of his reports on clinical cases \cite{mangasarian1995breast}. The data used has 369 samples, with 9 features listed below, and two outcome classes (benign and malignant). Features are scored on a scale of 1-10, and are (1) Clump Thickness, (2) Uniformity of Cell Size, (3) Uniformity of Cell Shape, (4) Marginal Adhesion, (5) Single Epithelial Cell Size, (6) Bare Nuclei, (7) Bland Chromatin, (8) Normal Nucleoli and (9) Mitoses.

We are particularly interested in providing an explanation, in addition to giving the right classification of outliers (malignant samples), as this can provide useful guidance for physicians to further investigate why certain cells are potentially harmful. We illustrate how \method attempts to explain its outlier detection process. For each dimension $j$  (9 dimensions in total) of the $i$-th data point, we plot $O(X_i^{(j)})$, the dimensional outlier score in black. Clearly, different dimensions show different contributions. We also plot the $99^{th}$ percentile band in green as a reference line
(which corresponds to $1 - \alpha$ percentage of outliers). In the analysis below, we plot the Dimensional Outlier Graph for 70-$th$ sample ($i=70$), which is malignant (outlier) and has been successfully classified by \method.

\begin{figure}[tp]\label{od70}
    \includegraphics[width=0.48\textwidth]{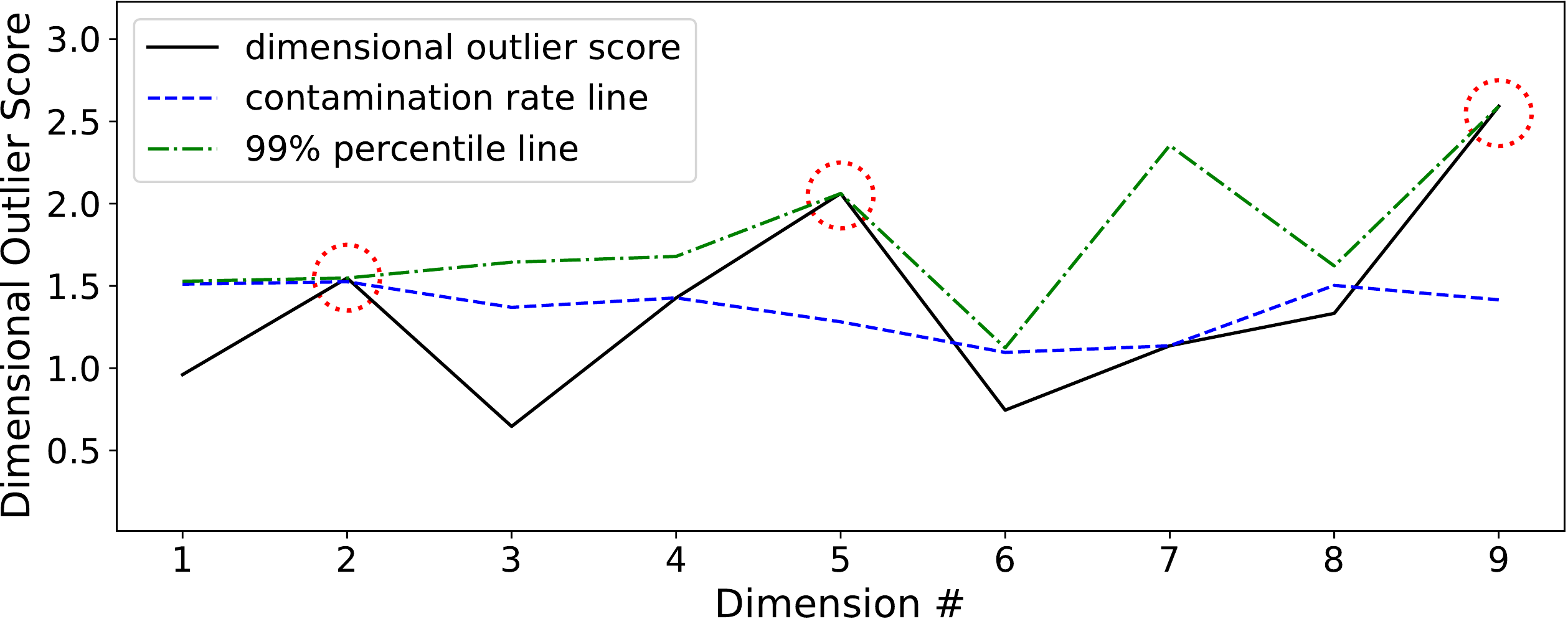}
    \caption{\textit{Dimensional Outlier Graph} for sample 70 (classified as an outlier) of BreastW dataset; dimension 2, 5, 9 are identified with most contribution to its outlyingness.}
\end{figure}

We see that for dimension/feature 2, 5, and 9, the dimensional outlier scores have ``touched" the $99^{th}$ percentile band, while all other features remain below the contamination rate band. This suggests that this point is likely an outlier because it is extremely non-uniform in size (dim. 2), has a large single epithelial cell size (dim. 5), and is more likely to reproduce via mitosis rather than meiosis (dim. 9).

\subsubsection{Complexity Analysis}
\label{subsubsec:complexity}
\method has $\mathcal{O}(nd)$ time and space complexity, and we break down the analysis by steps. In the estimation steps (Section \ref{subsubsec:rare_events}), computing ECDF for all $d$ dimensions using $n$ samples leads to $\mathcal{O}(nd)$ time and space complexity. In the aggregation steps (Section \ref{subsubsec:outlier_scores}), tail probability calculation and aggregation also lead to $\mathcal{O}(nd)$ time and space complexity.

\subsubsection{Acceleration by Distributed Learning}
\label{subsec:distributed}
Since \method estimates each dimension independently, it is suited for distributed learning acceleration with multiple workers. Given there are $t$ available workers for distributed computing, e.g., $t$ cores on a single machine. This constructs the worker pool as $\mathcal{W}=\{W_1,...,W_t\}$. Without distributed learning, \method will estimate each dimension $j \in \{1,\ldots,d\}$ of $\mathbf{X}$ iteratively, e.g., with a for loop. With multiple workers available ($t>1$), a generic scheduling system equally splits $d$ ECDF estimation and tail probability computation tasks into $t$ groups, so each available worker in $\mathcal{W}$ will process roughly $\lceil \frac{d}{t} \rceil $ estimation and computation task. Similar to the non-distributed setting, the results from each dimension will then be aggregated to generate the final outlier scores. We provide both single-thread and distributed implementations in PyOD \cite{zhao2019pyod}.

\newpage

\section{Empirical Evaluation}
\label{sec:empirical_evaluation}

\begin{table}[!htb]
\centering
	\caption{30 real-world benchmark datasets used in this study for evaluation. The datasets from the \texttt{ODDS} repo. are appended with ``(mat)", and the datasets from the \texttt{DAMI} repo. are appended with ``(arff)". For the \textit{Shuttle} dataset in both databases, we randomly subsample 10,000 samples.} % title of Table
	\footnotesize
	\begin{tabular}{l | rrr } % centered columns (4 columns)
		\toprule
		\textbf{Dataset} & \textbf{\# Samp. ($n$)} & \textbf{\# Dims. ($d$)} & \textbf{\% Outlier}\\
		\midrule
        Arrhythmia (mat) & 452 & 274 & 14.601 \\
        Breastw (mat) & 683 & 9 & 34.992 \\
        Cardio (mat) & 1831 & 21 & 9.612 \\
        Ionosphere (mat) & 351 & 33 & 35.897 \\
        Lympho (mat) & 148 & 18 & 4.054 \\
        Mammography (mat) & 11183 & 6 & 2.325 \\
        Optdigits (mat) & 5216 & 64 & 2.875 \\
        Pima (mat) & 768 & 8 & 34.895 \\
        Satellite (mat) & 6435 & 36 & 31.639 \\
        Satimage-2 (mat) & 5803 & 36 & 1.223 \\
        Shuttle (mat) (*) & 10000 & 9 & 7.120 \\
        Speech (mat) & 3686 & 400 & 1.654 \\
        WBC (mat) & 278 & 30 & 5.555 \\
        Wine (mat) & 129 & 13 & 7.751 \\
        \midrule
        Arrhythmia (arff) & 450 & 259 & 45.777 \\
        Cardiotocography (arff) & 2114 & 21 & 22.043 \\
        HeartDisease (arff) & 270 & 13 & 44.444 \\
        Hepatitis (arff) & 80 & 19 & 16.250 \\
        InternetAds (arff) & 1966 & 1555 & 18.718 \\
        Ionosphere (arff) & 768 & 8 & 34.895 \\
        KDDCup99 (arff) & 4207 & 57 & 39.909 \\
        Lymphography (arff) & 148 & 18 & 4.05 \\
        Pima (arff) & 768 & 8 & 34.90 \\
        Shuttle (arff) (*) & 10000 & 41 & 0.156 \\
        SpamBase (arff) & 4207 & 57 & 39.91 \\
        Stamps (arff) & 340 & 9 & 9.12 \\
        Waveform (arff) & 3443 & 21 & 2.904 \\
        WBC (arff) & 223 & 9 & 4.484 \\
        WDBC (arff) & 367 & 30 & 2.724 \\
        WPBC (arff) & 198 & 33 & 23.737 \\
		\bottomrule
	\end{tabular}
	\label{table:data} % is used to refer this table in the text
\end{table}
Our experiments answer the following questions:
\begin{enumerate}
    \item Among the variants of \method using different tail probabilities, which one yields the best performance? (\S \ref{subsec:variant_comparison})
    \item How effective is \method, in comparison to state-of-the-art (SOTA) outlier detectors? (\S \ref{subsubsec:overall})
    \item Under which conditions, the performance of \method may degrade? (\S \ref{subsubsec:case})
    \item How efficient and scalable is \method regarding high-dimensional, large datasets? (\S \ref{subsec:scalability})
\end{enumerate}

\subsection{Experiment Setup} 

\textbf{Datasets}. Table \ref{table:data} summarises 30 public outlier detection benchmark datasets used in this study from \texttt{ODDS}\footnote{ODDS Library: http://odds.cs.stonybrook.edu} \cite{Rayana:2016} and \texttt{DAMI}\footnote{DAMI Datasets: http://www.dbs.ifi.lmu.de/research/\\
outlier-evaluation/DAMI} \cite{dami} data repositories. Two special notes should be made about the datasets. First, multiple versions of the same dataset exist in different literature. For example, two versions of the \textit{Pima} dataset exist: the \textit{.mat} version has 8 dimensions and 768 observations, while the \textit{.arff} version has 32 dimensions and 351 observations. This is because different subsets of the original data were used by different authors, and they keep different dimensions and/or observations. For clarity, we specify the source of a dataset by appending ``(mat)" (from \texttt{ODDS}) or ``(arff)" (from \texttt{DAMI}) to its name. Second, because some baselines fail to converge for a large number of observations, we truncate both versions of \textit{Shuttle} down to 10,000 observations. The truncation is done by taking 10,000 random rows from the original datasets (which have slightly over 60,000 observations each).

\noindent \textbf{Evaluation metrics}. In each experiment, 60\% of the data is used for training and the remaining 40\% is set aside for testing. Performance is evaluated by taking the average score of 10 independent trials using the area under the receiver operating characteristic (ROC) and average precision (AP). Both metrics are widely used in outlier mining research \cite{Aggarwal2017,Akoglu2012,zhao2018xgbod,Emmott2015}. We report both the raw comparison results and the critical difference (CD) plots to show statistical difference \cite{demvsar2006statistical,fawaz2019deep}, where it visualizes the statistical comparison by Wilcoxon signed-rank test with Holm’s correction.

\noindent \textbf{Baselines, Implementation, and Environment}. We compare the performance of \method with 11 leading outlier detectors. We aim to include a variety of detectors to make the comparison robust. Specifically, the 11 competitors are Angle-Based Outlier Detection (ABOD) \cite{kriegel2008angle}, Clustering-Based Local Outlier Factor (CBLOF) \cite{he2003discovering},  Histogram-based Outlier Score (HBOS) \cite{goldstein2012histogram}, Isolation Forest (IForest) \cite{liu2008isolation}, k Nearest Neighbors (KNN) \cite{Ramaswamy2000efficient}, Lightweight On-line Detector of Anomalies (LODA) \cite{pevny2016loda}, Local Outlier Factor(LOF) \cite{Breunig2000lof}, Locally Selective Combination in Parallel Outlier Ensembles (LSCP) \cite{zhao2019lscp}, One-Class Support Vector Machines (OCSVM) \cite{scholkopf2001estimating}, PCA-based outlier detector (PCA) \cite{shyu2003novel}, and Scalable Unsupervised Outlier Detection (SUOD) \cite{zhao2021suod}. Their technical strength and limitations are discussed in Section \ref{sec:related_work}.

Notably, PyOD is a popular open-source Python toolbox for performing scalable outlier detection on multivariate data \cite{zhao2019pyod}. A wide range of outlier detection algorithms are included under a single, well-documented API.
This allows us to easily compare \method with other baseline detectors. Our implementation of \method is also under the framework of PyOD, and both single-process and distributed versions are readily available in PyOD\footnote{\method implementation in PyOD library: \url{https://github.com/yzhao062/pyod/blob/master/pyod/models/ecod.py}}. For a fair comparison, we use the single-process version in this study. In the subsequent experiments, a Windows laptop with Intel i5-8265U @ 1.60 GHz quad-core CPU and 8GB of memory are used.

\begin{figure*}[!htp] % [!htb]
\centering
\subfloat[\textit{Breastw (mat)}]{%
  \includegraphics[clip,width=0.5\columnwidth]{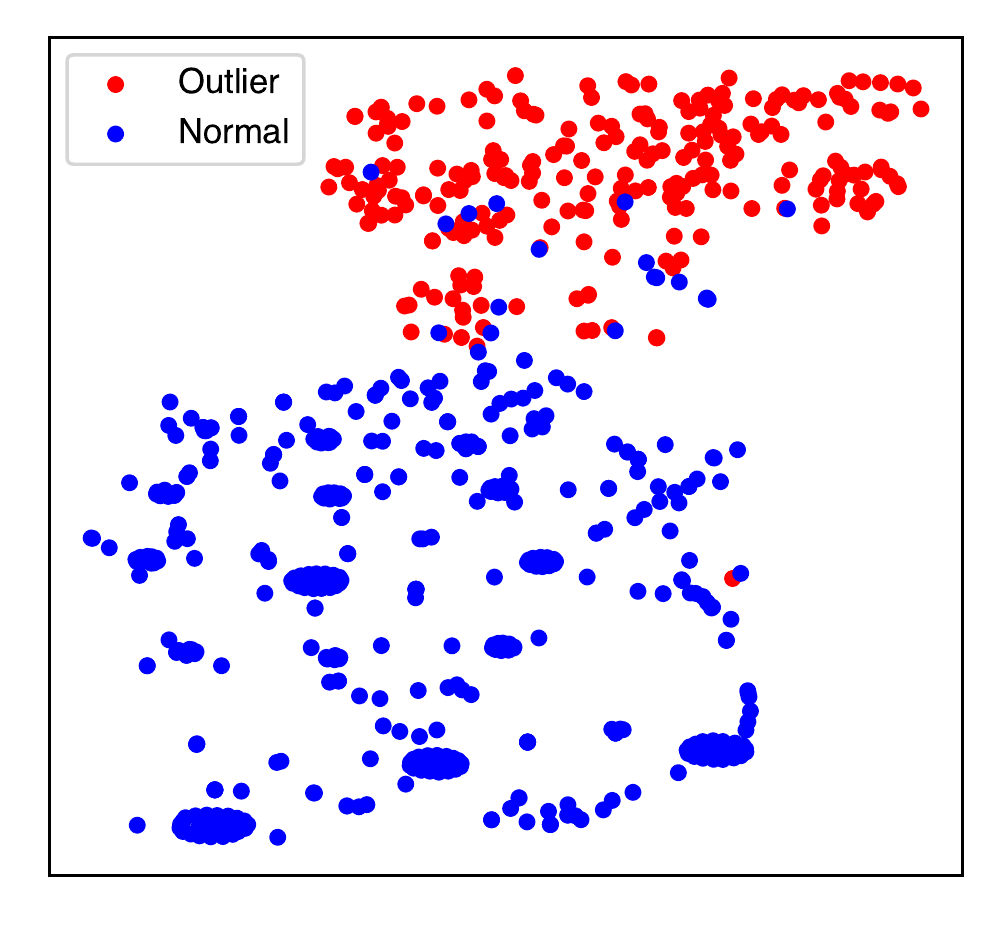}%
}
\subfloat[\textit{Shuttle (mat)}]{%
  \includegraphics[clip,width=0.5\columnwidth]{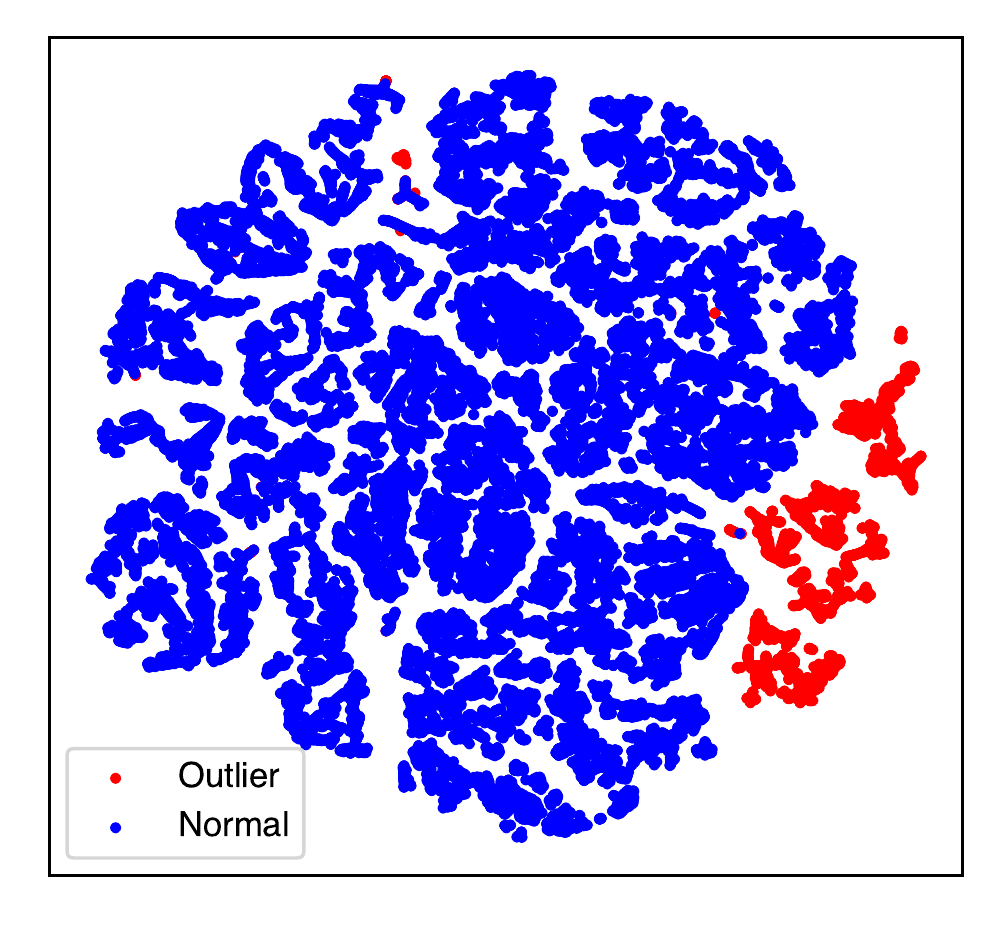}%
}
\subfloat[\textit{Ionosphere (mat)}]{%
  \includegraphics[clip,width=0.5\columnwidth]{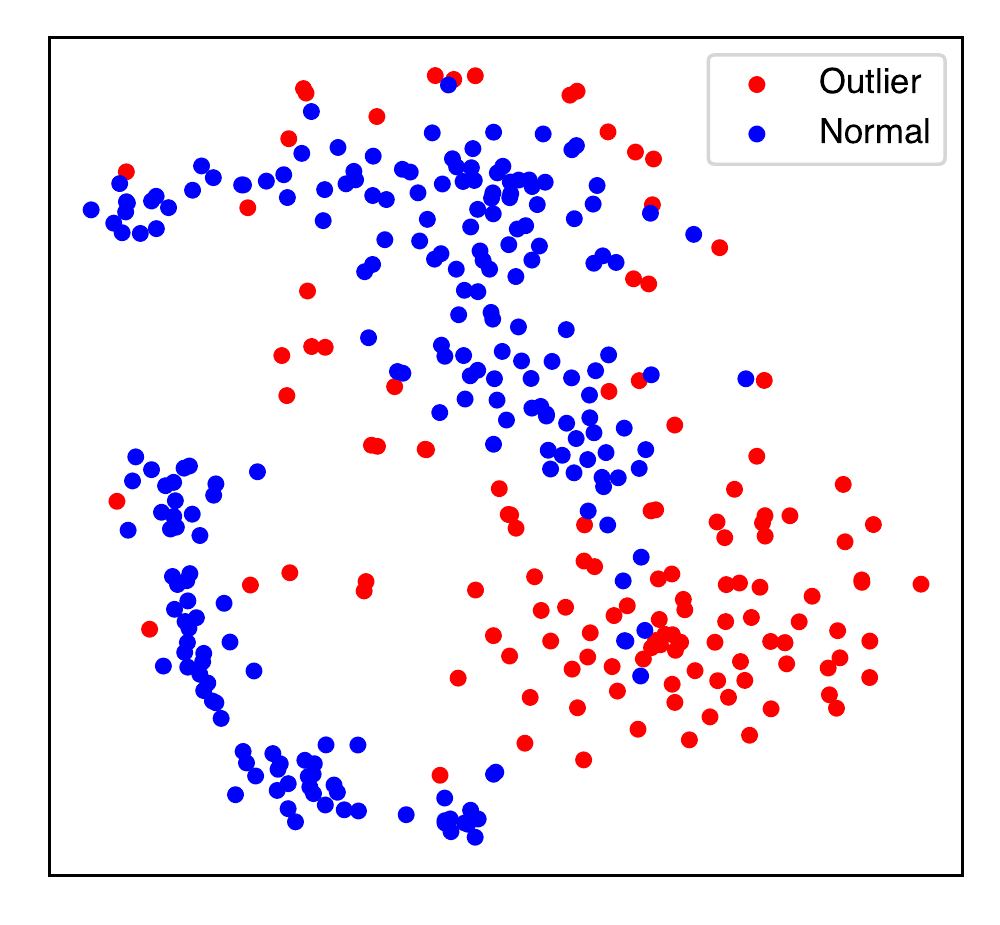}%
}
\subfloat[\textit{Speech (mat)}]{%
  \includegraphics[clip,width=0.5\columnwidth]{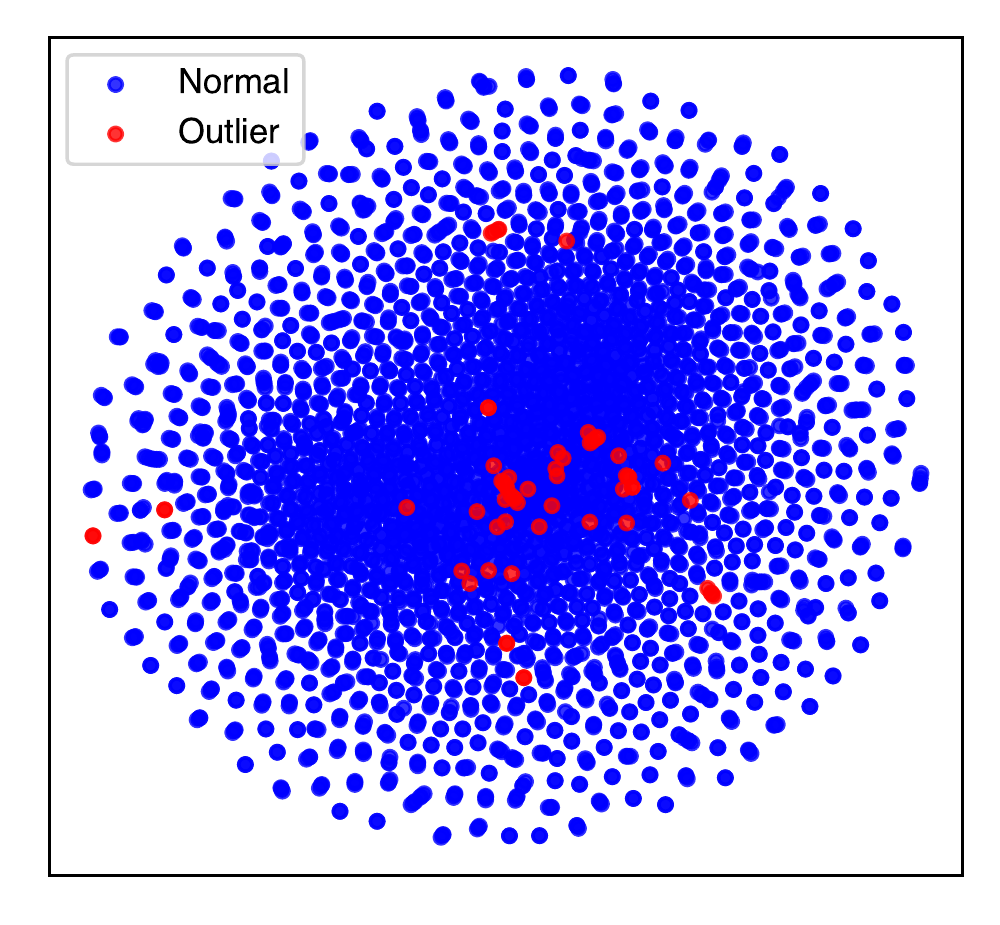}%
}
\\
\caption{2-D embedding of selected datasets. In the first two datasets (\textit{Breastw} and \textit{Shuttle}), the outliers locate at the tail for at least one dimension, and \method is the best algorithm in comparison to the baselines. In the last two datasets (\textit{Ionosphere} and \textit{Speech}), where outliers do not locate at the tails in any dimension but mingle with inliers, \method's performance degrades. }
\label{fig:case_study}
\end{figure*}

\begin{table}[!tp]
\setlength{\tabcolsep}{3pt}
\centering
	\caption{ROC score of 3 \method variants and \method (average of 10 independent trials, highest score highlighted in bold). Clearly, \method (the rightmost column) outperforms (ranking 1st in 18 out of 30 dataset).}
	\footnotesize
	\begin{tabular}{l|llll} % centered columns (4 columns)
    \toprule
    Data & \method-L & \method-R & \method-B & \method\\
    \midrule
    Arrhythmia (mat)        & 0.716 & 0.659          & 0.760          & \textbf{0.802} \\
Breastw (mat)           & 0.665 & 0.641          & 0.774          & \textbf{0.993} \\
Cardio (mat)            & 0.286 & 0.725          & 0.589          & \textbf{0.897} \\
Ionosphere (mat)        & 0.765 & 0.523          & 0.810          & \textbf{0.830} \\
Lympho (mat)            & 0.406 & 0.677          & 0.677          & \textbf{0.993} \\
Mammography (mat)       & 0.718 & 0.384          & 0.706          & \textbf{0.894} \\
Optdigits (mat)         & 0.993 & 0.997          & \textbf{0.997} & 0.732          \\
Pima (mat)              & 0.550 & 0.990          & \textbf{1.000}      & 0.663          \\
Satellite (mat)         & 0.279 & \textbf{0.736} & 0.604          & 0.661          \\
Satimage-2 (mat)        & 0.490 & 0.819          & 0.792          & \textbf{0.985} \\
Shuttle (mat)           & 0.151 & 0.707          & 0.664          & \textbf{0.990}  \\
Speech (mat)            & 0.688 & 0.746          & \textbf{0.873} & 0.484          \\
WBC (mat)               & 0.578 & 0.532          & 0.588          & \textbf{0.974} \\
Wine (mat)              & 0.025 & \textbf{0.990} & 0.988          & 0.949           \\
Arrhythmia (arff)       & 0.012 & \textbf{0.984} & 0.924          & 0.761          \\
Cardiotocography (arff) & 0.432 & 0.576          & 0.493          & \textbf{0.637}  \\
HeartDisease (arff)     & 0.805 & 0.687          & 0.821          & \textbf{0.672} \\
Hepatitis (arff)        & 0.008 & \textbf{0.993} & 0.991          & 0.843          \\
InternetAds (arff)      & 0.887 & 0.687          & \textbf{0.936} & 0.676          \\
Ionosphere (arff)       & 0.970 & 0.316          & \textbf{0.933} & 0.821          \\
KDDCup99 (arff)         & 0.730 & 0.383          & 0.715          & \textbf{0.997} \\
Lymphography (arff)     & 0.884 & 0.732          & 0.995          & \textbf{0.998} \\
Pima (arff)             & 0.241 & 0.892          & \textbf{0.898} & 0.646          \\
Shuttle (arff)          & 0.460 & 0.666          & 0.693          & \textbf{0.879} \\
SpamBase (arff)         & 0.225 & \textbf{0.755} & 0.584          & 0.700          \\
Stamps (arff)           & 0.749 & 0.414          & 0.581          & \textbf{0.938} \\
Waveform (arff)         & 0.926 & 0.813          & \textbf{0.973} & 0.786          \\
WBC (arff)              & 0.461  & 0.458          & 0.450          & \textbf{0.990} \\
WDBC (arff)             & 0.031 & 0.975          & 0.884           & \textbf{0.982} \\
WPBC (arff)             & 0.040 & \textbf{0.965} & 0.748           & 0.546          \\
\midrule
AVG                     & 0.560 & 0.714          & 0.781          & \textbf{0.824} \\

\bottomrule
	\end{tabular}
	\label{table:ECDO comparison ROC} % is used to refer this table in the text
\end{table}
\begin{table}[!htp]
\setlength{\tabcolsep}{3pt}
\centering
	\caption{Average precision (AP) of 3 \method variants and \method (average of 10 independent trials, highest score highlighted in bold). Clearly, \method (the rightmost column) outperforms (ranking 1st in 18 out of 30 dataset).}
	\footnotesize
	\begin{tabular}{l|llll} % centered columns (4 columns)
    \toprule
%  & \multicolumn{4}{c}{ROC-AUC} & \multicolumn{4}{c}{Average Precision} \\
%     \midrule
    Data & \method-L & \method-R & \method-B & \method\\
    \midrule
    Arrhythmia (mat)        & \textbf{0.503} & 0.318          & 0.497          & 0.472          \\
Breastw (mat)           & 0.209         & 0.987          & 0.984          & \textbf{0.987} \\
Cardio (mat)            & 0.450            & 0.163           & \textbf{0.587} & 0.579          \\
Ionosphere (mat)        & 0.363          & 0.006          & 0.151          & \textbf{0.719} \\
Lympho (mat)            & 0.592          & 0.334          & 0.622          & \textbf{0.893} \\
Mammography (mat)       & 0.528          & 0.431          & \textbf{0.912} & 0.429           \\
Optdigits (mat)         & 0.013          & 0.392          & \textbf{0.426} & 0.053          \\
Pima (mat)              & 0.026          & 0.049           & 0.045         & \textbf{0.540} \\
Satellite (mat)         & 0.229          & \textbf{0.611} & 0.460          & 0.585          \\
Satimage-2 (mat)        & 0.550          & 0.259          & 0.523          & \textbf{0.859} \\
Shuttle (mat)           & 0.113          & 0.044          & 0.670          & \textbf{0.980} \\
Speech (mat)            & 0.015          & 0.022          & \textbf{0.022} & 0.019          \\
WBC (mat)               & 0.035          & 0.772          & 0.464          & \textbf{0.782} \\
Wine (mat)              & 0.044          & \textbf{0.666} & 0.218           & 0.608          \\
Arrhythmia (arff)       & 0.710          & 0.636           & 0.751          & \textbf{0.751} \\
Cardiotocography (arff) & 0.370          & 0.294          & \textbf{0.493} & 0.377          \\
HeartDisease (arff)     & 0.319          & \textbf{0.663} & 0.508          & 0.640            \\
Hepatitis (arff)        & 0.444          & 0.221          & 0.434          & \textbf{0.584} \\
InternetAds (arff)      & 0.156          & 0.511          & \textbf{0.512} & 0.510          \\
Ionosphere (arff)       & 0.572          & 0.334          & \textbf{0.607} & 0.530          \\
KDDCup99 (arff)         & 0.166          & 0.219          & 0.278          & \textbf{0.566} \\
Lymphography (arff)     & 0.120          & 0.986         & \textbf{1.000}      & 0.463          \\
Pima (arff)             & 0.244          & 0.590          & 0.477          & \textbf{0.703} \\
Shuttle (arff)          & 0.014          & 0.068         & 0.073          & \textbf{0.238} \\
SpamBase (arff)         & 0.254          & 0.568          & 0.527          & \textbf{0.981} \\
Stamps (arff)           & 0.199          & 0.220          & \textbf{0.352} & 0.085          \\
Waveform (arff)         & 0.049          & 0.031          & 0.042          & \textbf{0.078} \\
WBC (arff)              & 0.028          & 0.838          & 0.820          & \textbf{0.838} \\
WDBC (arff)             & 0.015          & \textbf{0.842} & 0.622           & 0.840          \\
WPBC (arff)             & 0.199          & \textbf{0.266} & 0.211          & 0.243          \\
\midrule
AVG                     & 0.251          & 0.411 & 0.476          & \textbf{0.564}  \\

\bottomrule
	\end{tabular}
	\label{table:ECDO comparison AP} % is used to refer this table in the text
\end{table}

\subsection{The Effect of Using Different Tails with \method}
\label{subsec:variant_comparison}
How does the usage of different tail probabilities affect the performance of \method? In the first experiment, we compare the performances of \method while using different tail probabilities for measuring sample outlyingness. Specifically, the three variants and \method are compared: (1) only using left tail probability (\method-L) (2) only using right tail probability (\method-R) (3) using the average of both tail probabilities (\method-B) and (4) using automatically selected tail probability (\method) as outlined in Section  \ref{subsubsec:outlier_scores}. Note that with the three  variants, the outlier score of the $i$-th sample is equal to $O_{\textsf{left-only}}(X_i)$, $O_{\textsf{right}}(X_i)$, and $\frac{1}{2}(O_{\textsf{left-only}}(X_i)+O_{\textsf{right-only}}(X_i))$ respectively, where $O_{\textsf{left-only}}(X_i)$ and $O_{\textsf{right-only}}(X_i)$ are the sum of negative log probabilities defined in Eq.~(\ref{eq:right_only}). Differently, \method uses both tails automatically as described in Eq. (\ref{eq:outlier_scores})
Table \ref{table:ECDO comparison ROC} and \ref{table:ECDO comparison AP} show the results comparison regarding both ROC and AP on the benchmark datasets. 

\textbf{The proposed \method achieves the best performance}, with an average ROC of 0.824 and average precision of 0.564. It is better than \method-B using the average of two tails (avg. ROC: 0.781, avg. AP: 0.476), \method-R using right tails (avg. ROC: 0.714, avg. AP: 0.411), and \method-L using left tails (avg. ROC: 0.506, avg. AP: 0.251). Its superiority can be credited to the automatic selection of tail probabilities by the skewness of the underlying datasets. Clearly, the 30 benchmark datasets would have outliers on the different tails of the distribution, and none of the variants can capture this, although taking the average in \method-B alleviates the problem and therefore ranks at the second position. Consequently, we use the automatic version in \method by default, which is more carefully analyzed throughout the paper.

\subsection{Performance Comparison with Baselines}
\label{subsec:performance_comparision}
\begin{figure}[!tb] % [!htb]
\centering
\subfloat[Critical difference diagram of ROC]{%
  \includegraphics[clip,width=\columnwidth]{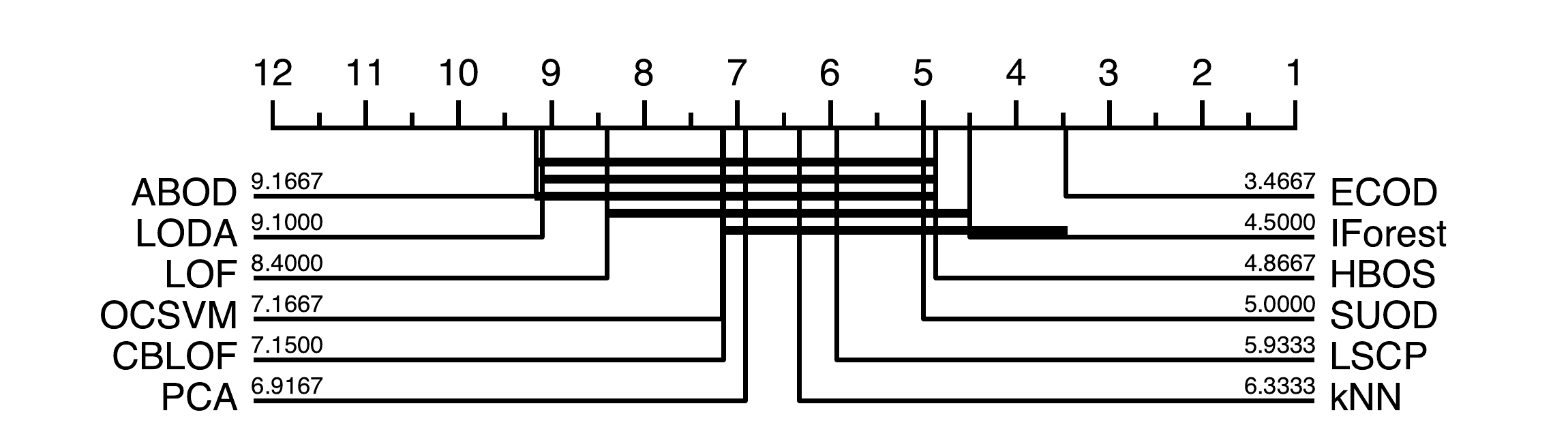}%
}

\subfloat[Critical difference diagram of AP]{%
  \includegraphics[clip,width=\columnwidth]{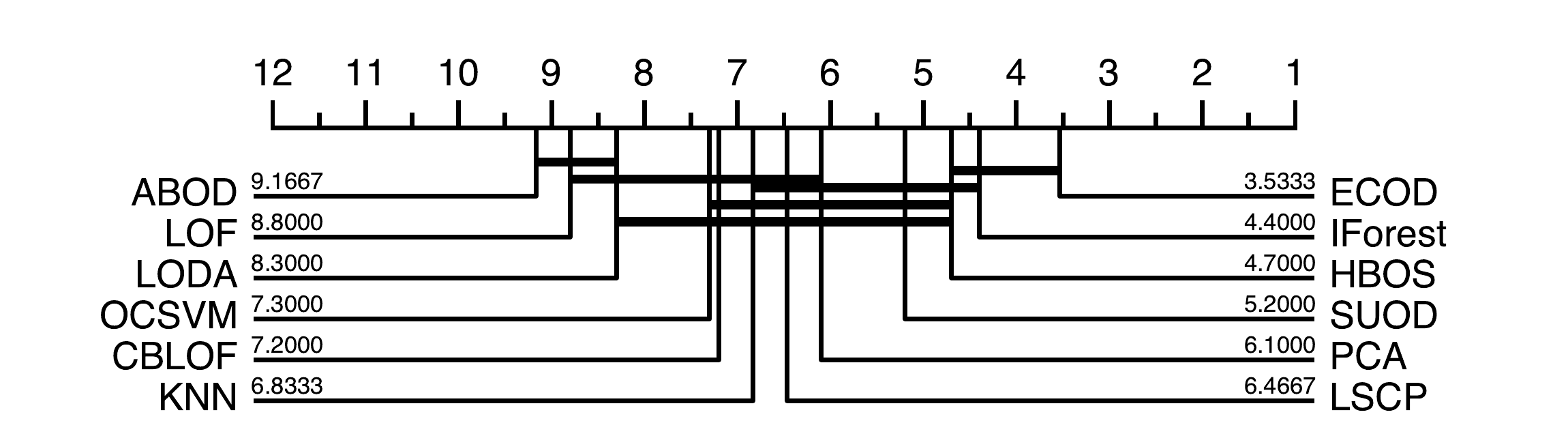}%
}
\\
\caption{Critical difference diagram showing pairwise statistical difference comparison of \method and the baselines regarding both ROC and AP. \method outperforms all baselines, and the result is statistically significant regarding ROC. }
\label{fig:cd_plot}
\end{figure}
\begin{table*}[!ht]
\setlength{\tabcolsep}{3pt}
\centering
	\caption{ROC scores of detector performance (average of 10 independent trials, highest score highlighted in bold); rank is shown in parenthesis (lower is better). \method outperforms all baselines with the highest avg. score and ranks the highest in 13 out of 30 datasets. }
\vspace{-0.1in}

\scalebox{1}{
\scriptsize
\begin{tabular}{l|lllllllllll|l}
\toprule
\textbf{Data}           & \textbf{ABOD}      & \textbf{CBLOF}     & \textbf{HBOS}      & \textbf{IForest}   & \textbf{KNN}       & \textbf{LODA} & \textbf{LOF}       & \textbf{LSCP} & \textbf{OCSVM} & \textbf{PCA}       & \textbf{SUOD}      & \textbf{ECOD}      \\
\midrule
Arrhythmia (mat)        & 0.769 (11)         & 0.784 (7)          & \textbf{0.822 (1)} & 0.802 (3)          & 0.786 (6)          & 0.75 (12)     & 0.779 (10)         & 0.793 (5)     & 0.781 (9)      & 0.782 (8)          & 0.811 (2)          & 0.802 (3)          \\
Breastw (mat)           & 0.396 (12)         & 0.965 (6)          & 0.983 (4)          & 0.987 (2)          & 0.976 (5)          & 0.987 (2)     & 0.47 (11)          & 0.933 (9)     & 0.961 (7)      & 0.959 (8)          & 0.929 (10)         & \textbf{0.994 (1)} \\
Cardio (mat)            & 0.569 (12)         & 0.81 (9)           & 0.835 (8)          & 0.923 (3)          & 0.724 (10)         & 0.856 (7)     & 0.574 (11)         & 0.88 (6)      & 0.935 (2)      & \textbf{0.95 (1)}  & 0.9 (4)            & 0.897 (5)          \\
Ionosphere (mat)        & 0.925 (2)          & 0.897 (3)          & 0.561 (12)         & 0.847 (7)          & \textbf{0.927 (1)} & 0.798 (10)    & 0.875 (4)          & 0.857 (6)     & 0.842 (8)      & 0.796 (11)         & 0.862 (5)          & 0.831 (9)          \\
Lympho (mat)            & 0.911 (11)         & 0.967 (10)         & \textbf{0.996 (1)} & 0.993 (3)          & 0.975 (9)          & 0.73 (12)     & 0.977 (7)          & 0.984 (5)     & 0.976 (8)      & 0.985 (4)          & 0.979 (6)          & 0.994 (2)          \\
Mammography (mat)       & 0.549 (12)         & 0.821 (10)         & 0.834 (9)          & 0.86 (5)           & 0.841 (8)          & 0.881 (3)     & 0.729 (11)         & 0.855 (6)     & 0.872 (4)      & 0.882 (2)          & 0.851 (7)          & \textbf{0.894 (1)} \\
Optdigits (mat)         & 0.467 (9)          & 0.769 (2)          & \textbf{0.873 (1)} & 0.721 (4)          & 0.371 (12)         & 0.398 (11)    & 0.45 (10)          & 0.658 (6)     & 0.5 (8)        & 0.509 (7)          & 0.68 (5)           & 0.733 (3)          \\
Pima (mat)              & 0.679 (3)          & 0.658 (8)          & 0.7 (2)            & 0.678 (4)          & \textbf{0.708 (1)} & 0.592 (12)    & 0.627 (10)         & 0.664 (6)     & 0.622 (11)     & 0.648 (9)          & 0.666 (5)          & 0.664 (6)          \\
Satellite (mat)         & 0.571 (11)         & 0.749 (2)          & \textbf{0.758 (1)} & 0.702 (4)          & 0.684 (5)          & 0.597 (10)    & 0.557 (12)         & 0.665 (6)     & 0.662 (7)      & 0.599 (9)          & 0.722 (3)          & 0.661 (8)          \\
Satimage-2 (mat)        & 0.819 (11)         & \textbf{0.999 (1)} & 0.98 (9)           & 0.995 (3)          & 0.954 (10)         & 0.987 (5)     & 0.458 (12)         & 0.981 (8)     & 0.998 (2)      & 0.982 (7)          & 0.993 (4)          & 0.985 (6)          \\
Shuttle (mat)           & 0.659 (9)          & 0.608 (10)         & 0.995 (4)          & \textbf{0.998 (1)} & 0.738 (8)          & 0.589 (11)    & 0.524 (12)         & 0.939 (7)     & 0.995 (4)      & 0.994 (6)          & 0.996 (3)          & \textbf{0.998 (1)} \\
Speech (mat)            & \textbf{0.627 (1)} & 0.452 (9)          & 0.454 (8)          & 0.447 (11)         & 0.458 (6)          & 0.455 (7)     & 0.471 (3)          & 0.465 (5)     & 0.447 (11)     & 0.45 (10)          & 0.469 (4)          & 0.485 (2)          \\
Wbc (mat)               & 0.905 (12)         & 0.92 (9)           & 0.952 (2)          & 0.931 (7)          & 0.937 (3)          & 0.913 (11)    & 0.935 (4)          & 0.931 (7)     & 0.932 (6)      & 0.916 (10)         & 0.935 (4)          & \textbf{0.975 (1)} \\
Wine (mat)              & 0.431 (11)         & 0.284 (12)         & 0.9 (3)            & 0.816 (6)          & 0.518 (10)         & 0.787 (8)     & 0.905 (2)          & 0.862 (5)     & 0.636 (9)      & 0.801 (7)          & 0.868 (4)          & \textbf{0.949 (1)} \\
Arrhythmia (arff)       & 0.74 (11)          & 0.751 (6)          & 0.749 (9)          & 0.757 (3)          & 0.751 (6)          & 0.703 (12)    & 0.749 (9)          & 0.757 (3)     & 0.756 (5)      & 0.751 (6)          & \textbf{0.773 (1)} & 0.762 (2)          \\
Cardiotocography (arff) & 0.459 (12)         & 0.567 (9)          & 0.594 (8)          & 0.683 (3)          & 0.49 (11)          & 0.674 (4)     & 0.505 (10)         & 0.645 (6)     & 0.698 (2)      & \textbf{0.755 (1)} & 0.66 (5)           & 0.637 (7)          \\
HeartDisease (arff)     & 0.597 (6)          & 0.593 (7)          & \textbf{0.689 (1)} & 0.602 (4)          & 0.614 (3)          & 0.482 (12)    & 0.563 (10)         & 0.599 (5)     & 0.555 (11)     & 0.576 (9)          & 0.587 (8)          & 0.673 (2)          \\
Hepatitis (arff)        & 0.716 (11)         & 0.759 (10)         & 0.796 (6)          & 0.794 (7)          & 0.811 (4)          & 0.608 (12)    & 0.831 (2)          & 0.784 (8)     & 0.764 (9)      & 0.804 (5)          & 0.814 (3)          & \textbf{0.843 (1)} \\
InternetAds (arff)      & 0.643 (6)          & 0.617 (10)         & \textbf{0.699 (1)} & 0.685 (2)          & 0.622 (8)          & 0.528 (12)    & 0.606 (11)         & 0.675 (4)     & 0.623 (7)      & 0.622 (8)          & 0.673 (5)          & 0.676 (3)          \\
Ionosphere (arff)       & \textbf{0.925 (1)} & 0.886 (3)          & 0.552 (12)         & 0.836 (8)          & 0.919 (2)          & 0.822 (9)     & 0.86 (5)           & 0.855 (6)     & 0.838 (7)      & 0.787 (11)         & 0.862 (4)          & 0.822 (9)          \\
KDDCup99 (arff)         & 0.693 (11)         & \textbf{0.997 (1)} & 0.995 (5)          & \textbf{0.997 (1)} & 0.761 (9)          & 0.728 (10)    & 0.577 (12)         & 0.96 (8)      & 0.995 (5)      & \textbf{0.997 (1)} & 0.995 (5)          & \textbf{0.997 (1)} \\
Lymphography (arff)     & 0.986 (11)         & 0.996 (7)          & 0.998 (3)          & 0.999 (1)          & 0.997 (4)          & 0.839 (12)    & 0.995 (8)          & 0.995 (8)     & 0.992 (10)     & 0.997 (4)          & 0.997 (4)          & \textbf{0.999 (1)} \\
Pima (arff)             & 0.667 (3)          & 0.666 (4)          & 0.659 (6)          & 0.676 (2)          & \textbf{0.706 (1)} & 0.597 (12)    & 0.658 (7)          & 0.661 (5)     & 0.64 (11)      & 0.655 (8)          & 0.65 (9)           & 0.647 (10)         \\
Shuttle (arff)          & 0.834 (10)         & 0.98 (2)           & 0.792 (11)         & 0.866 (9)          & 0.961 (4)          & 0.707 (12)    & \textbf{0.986 (1)} & 0.941 (6)     & 0.966 (3)      & 0.929 (7)          & 0.955 (5)          & 0.88 (8)           \\
SpamBase (arff)         & 0.435 (11)         & 0.548 (7)          & 0.67 (2)           & 0.633 (3)          & 0.54 (8)           & 0.445 (10)    & 0.426 (12)         & 0.588 (4)     & 0.539 (9)      & 0.555 (6)          & 0.562 (5)          & \textbf{0.701 (1)} \\
Stamps (arff)           & 0.762 (10)         & 0.73 (12)          & 0.918 (2)          & 0.913 (4)          & 0.874 (9)          & 0.901 (6)     & 0.739 (11)         & 0.906 (5)     & 0.88 (8)       & 0.917 (3)          & 0.898 (7)          & \textbf{0.939 (1)} \\
Waveform (arff)         & 0.652 (10)         & 0.714 (5)          & 0.683 (7)          & 0.681 (8)          & 0.74 (3)           & 0.622 (12)    & 0.735 (4)          & 0.76 (2)      & 0.665 (9)      & 0.634 (11)         & 0.706 (6)          & \textbf{0.787 (1)} \\
WBC (arff)              & 0.956 (11)         & 0.966 (10)         & 0.982 (3)          & \textbf{0.991 (1)} & 0.972 (8)          & 0.981 (4)     & 0.932 (12)         & 0.967 (9)     & 0.975 (6)      & 0.974 (7)          & 0.981 (4)          & 0.99 (2)           \\
WDBC (arff)             & 0.903 (12)         & 0.909 (11)         & 0.967 (2)          & 0.946 (3)          & 0.927 (9)          & 0.946 (3)     & 0.92 (10)          & 0.937 (6)     & 0.938 (5)      & 0.929 (8)          & 0.932 (7)          & \textbf{0.983 (1)} \\
WPBC (arff)             & 0.457 (12)         & 0.492 (11)         & 0.536 (2)          & 0.516 (6)          & 0.526 (3)          & 0.503 (8)     & 0.511 (7)          & 0.524 (5)     & 0.498 (9)      & 0.496 (10)         & 0.526 (3)          & \textbf{0.547 (1)} \\
\midrule
AVG                     & 0.69 (12)          & 0.762 (8)          & 0.797 (5)          & 0.809 (2)          & 0.76 (9)           & 0.713 (10)    & 0.697 (11)         & 0.801 (4)     & 0.783 (7)      & 0.788 (6)          & 0.808 (3)          & \textbf{0.825 (1)} \\
\bottomrule
\end{tabular}
}
	\label{table:od comparison roc} % is used to refer this table in the text
\end{table*}

\subsubsection{Overall Results}
\label{subsubsec:overall}
\textbf{\method consistently outperforms regarding both ROC and AP}. Of the 30 datasets in Table \ref{table:data}, \method scores the highest in terms of ROC (see Table \ref{table:od comparison roc}). It achieves an average ROC score of 0.825, which is 2\% higher than the second best alternative---iForest. Notably, iForest has been the SOTA method in many large-scale outlier analyses \cite{Emmott2015,ma2021large}. Moreover, \method ranks first in 13 out of 30 occasions and ranks in the top three places in 21 out of 30 occasions. The critical difference plot in Fig. \ref{fig:cd_plot}  corroborates the findings---it shows that \method outperforms with a statistical significance.

Similarly, Table \ref{table:od comparison map} shows that \method is also superior to the baselines regarding AP. Of the 12 OD methods, \method achieves an average AP score of 0.565, which bring 5\% improvement over the second place---iForest. Additionally, \method ranks first in 12 out of 30 datasets, and ranks in the top three places on 20 datasets. The further analysis with the critical difference plot in Fig. \ref{fig:cd_plot} corroborates that \method outperforms the baselines.

Additionally, it is worthy to note that \method (avg. ROC: 0.825, avg. AP: 0.565) is better than HBOS (avg. ROC: 0.795, avg. AP: 0.509) with a similar mechanism but estimates histogram per dimension instead. We credit the edge to: (1) \method uses more information than HBOS---the latter does not capture the difference for the samples in the same bin and (2) HBOS needs to decide the number of bins, which is hard to tune under unsupervised settings.

\subsubsection{Case Study}
\label{subsubsec:case}
Although \method outperforms on most of the datasets, we notice that its performance degrades markedly on some of the datasets, as shown in Table \ref{table:od comparison roc} and \ref{table:od comparison map}.
In the case study, we further investigate by visualizing the selected datasets in 2-D by t-SNE \cite{van2008visualizing}. 

In Fig. \ref{fig:case_study}, we first present two datasets (\textit{Breastw (mat)} and \textit{Shuttle (mat)}) where \method outperforms all baselines, and then two datasets (\textit{Ionosphere (mat)} and \textit{Speech (mat)}) where it does not achieve top performance. The visualization suggests that when outliers locate at the tails in at least some dimensions, \method could accurately capture them. However, its performance degrades when outliers are well mingled with normal points (Fig. \ref{fig:case_study}, subfigure c) or hidden in the middle of normal points regarding all dimensions (Fig. \ref{fig:case_study}, subfigure d). We want to point out that it is unlikely that an outlier resembles normal points in all dimensions, and 
this explains why \method could consistently work for most of the datasets.

\begin{table*}[!ht]
\setlength{\tabcolsep}{3pt}
\centering
	\caption{Average precision (AP) of detector performance (average of 10 independent trials, highest score highlighted in bold); rank is shown in parenthesis (lower is better). \method outperforms all baselines with the highest avg. performance, and ranks highest in 11 out of 30 datasets.}
\footnotesize
\scalebox{1}{
\scriptsize
\begin{tabular}{l|lllllllllll|l}
\toprule
\textbf{Data}           & \textbf{ABOD}      & \textbf{CBLOF}     & \textbf{HBOS}      & \textbf{IForest}   & \textbf{KNN}       & \textbf{LODA} & \textbf{LOF}       & \textbf{LSCP} & \textbf{OCSVM} & \textbf{PCA}       & \textbf{SUOD}      & \textbf{ECOD}      \\
\midrule
Arrhythmia (mat)        & 0.359 (12)         & 0.399 (9)          & \textbf{0.493 (3)} & \textbf{0.506 (1)} & 0.397 (10)         & 0.436 (5)     & 0.374 (11)         & 0.411 (6)          & 0.405 (7)      & 0.402 (8)          & 0.495 (2)          & 0.473 (4)          \\
Breastw (mat)           & 0.295 (12)         & 0.914 (8)          & 0.953 (5)          & 0.972 (3)          & 0.927 (7)          & 0.978 (2)     & 0.322 (11)         & 0.826 (9)          & 0.934 (6)      & 0.96 (4)           & 0.791 (10)         & \textbf{0.988 (1)} \\
Cardio (mat)            & 0.194 (11)         & 0.414 (8)          & 0.46 (6)           & 0.576 (3)          & 0.345 (10)         & 0.424 (7)     & 0.163 (12)         & 0.396 (9)          & 0.532 (4)      & \textbf{0.611 (1)} & 0.462 (5)          & 0.579 (2)          \\
Ionosphere (mat)        & 0.914 (2)          & 0.871 (3)          & 0.366 (12)         & 0.789 (8)          & \textbf{0.924 (1)} & 0.716 (11)    & 0.821 (6)          & 0.818 (7)          & 0.823 (5)      & 0.736 (9)          & 0.824 (4)          & 0.719 (10)         \\
Lympho (mat)            & 0.517 (11)         & 0.808 (9)          & \textbf{0.925 (2)} & \textbf{0.952 (1)} & 0.82 (7)           & 0.417 (12)    & 0.825 (6)          & 0.861 (4)          & 0.814 (8)      & 0.852 (5)          & 0.799 (10)         & 0.894 (3)          \\
Mammography (mat)       & 0.023 (12)         & 0.137 (9)          & 0.122 (10)         & 0.233 (3)          & 0.17 (6)           & 0.281 (2)     & 0.118 (11)         & 0.169 (7)          & 0.188 (5)      & 0.199 (4)          & 0.164 (8)          & \textbf{0.429 (1)} \\
Optdigits (mat)         & 0.028 (8)          & 0.062 (2)          & \textbf{0.196 (1)} & 0.055 (3)          & 0.022 (12)         & 0.024 (11)    & 0.029 (7)          & 0.043 (6)          & 0.028 (8)      & 0.028 (8)          & 0.046 (5)          & 0.053 (4)          \\
Pima (mat)              & 0.511 (4)          & 0.477 (7)          & \textbf{0.569 (1)} & 0.503 (5)          & \textbf{0.515 (3)} & 0.408 (12)    & 0.43 (11)          & 0.47 (9)           & 0.461 (10)     & 0.478 (6)          & 0.473 (8)          & 0.541 (2)          \\
Satellite (mat)         & 0.397 (11)         & \textbf{0.69 (1)}  & \textbf{0.687 (2)} & 0.654 (3)          & 0.543 (9)          & 0.581 (8)     & 0.39 (12)          & 0.51 (10)          & 0.653 (4)      & 0.603 (6)          & 0.608 (5)          & 0.585 (7)          \\
Satimage-2 (mat)        & 0.187 (11)         & \textbf{0.978 (1)} & 0.758 (7)          & 0.929 (3)          & 0.419 (9)          & 0.87 (5)      & 0.027 (12)         & 0.333 (10)         & 0.975 (2)      & 0.874 (4)          & 0.623 (8)          & 0.86 (6)           \\
Shuttle (mat)           & 0.171 (11)         & 0.195 (10)         & 0.98 (3)           & \textbf{0.986 (1)} & 0.204 (9)          & 0.379 (8)     & 0.142 (12)         & 0.715 (7)          & 0.902 (6)      & 0.926 (4)          & 0.913 (5)          & \textbf{0.981 (2)} \\
Speech (mat)            & \textbf{0.04 (1)}  & 0.022 (6)          & 0.027 (3)          & 0.018 (11)         & 0.022 (6)          & 0.017 (12)    & 0.024 (5)          & 0.027 (3)          & 0.021 (9)      & 0.022 (6)          & 0.029 (2)          & 0.02 (10)          \\
Wbc (mat)               & 0.355 (12)         & 0.5 (11)           & 0.663 (2)          & 0.59 (4)           & 0.529 (9)          & 0.564 (5)     & 0.558 (6)          & 0.554 (7)          & 0.514 (10)     & 0.534 (8)          & 0.602 (3)          & \textbf{0.783 (1)} \\
Wine (mat)              & 0.084 (11)         & 0.06 (12)          & 0.405 (2)          & 0.279 (6)          & 0.095 (10)         & 0.278 (7)     & 0.361 (4)          & 0.299 (5)          & 0.141 (9)      & 0.254 (8)          & 0.364 (3)          & \textbf{0.608 (1)} \\
Arrhythmia (arff)       & 0.699 (11)         & 0.712 (8)          & 0.75 (3)           & 0.746 (4)          & 0.712 (8)          & 0.697 (12)    & 0.704 (10)         & 0.718 (5)          & 0.716 (6)      & 0.714 (7)          & \textbf{0.751 (2)} & \textbf{0.752 (1)} \\
Cardiotocography (arff) & 0.247 (12)         & 0.363 (9)          & 0.366 (8)          & 0.434 (2)          & 0.311 (10)         & 0.432 (3)     & 0.258 (11)         & 0.374 (7)          & 0.419 (4)      & \textbf{0.478 (1)} & 0.398 (5)          & 0.378 (6)          \\
HeartDisease (arff)     & 0.534 (4)          & 0.521 (9)          & \textbf{0.625 (2)} & 0.534 (4)          & 0.538 (3)          & 0.445 (12)    & 0.478 (11)         & 0.525 (8)          & 0.513 (10)     & 0.53 (6)           & 0.528 (7)          & \textbf{0.64 (1)}  \\
Hepatitis (arff)        & 0.33 (12)          & 0.374 (11)         & 0.473 (6)          & 0.442 (8)          & 0.475 (5)          & 0.396 (10)    & 0.499 (4)          & 0.472 (7)          & 0.427 (9)      & 0.543 (3)          & 0.557 (2)          & \textbf{0.585 (1)} \\
InternetAds (arff)      & 0.276 (10)         & 0.315 (8)          & \textbf{0.535 (1)} & 0.49 (3)           & 0.281 (9)          & 0.242 (12)    & 0.262 (11)         & 0.387 (5)          & 0.316 (7)      & 0.32 (6)           & 0.475 (4)          & 0.51 (2)           \\
Ionosphere (arff)       & \textbf{0.915 (1)} & 0.862 (3)          & 0.364 (12)         & 0.777 (8)          & \textbf{0.915 (1)} & 0.763 (9)     & 0.811 (7)          & 0.819 (6)          & 0.822 (5)      & 0.723 (10)         & 0.835 (4)          & 0.703 (11)         \\
KDDCup99 (arff)         & 0.018 (12)         & \textbf{0.198 (5)} & \textbf{0.278 (1)} & \textbf{0.273 (2)} & 0.046 (10)         & 0.135 (7)     & 0.028 (11)         & 0.127 (8)          & 0.125 (9)      & \textbf{0.199 (4)} & 0.157 (6)          & \textbf{0.239 (3)} \\
Lymphography (arff)     & 0.801 (11)         & 0.925 (7)          & 0.978 (3)          & \textbf{0.992 (1)} & 0.942 (6)          & 0.491 (12)    & 0.925 (7)          & 0.922 (9)          & 0.837 (10)     & 0.953 (4)          & 0.944 (5)          & \textbf{0.982 (2)} \\
Pima (arff)             & 0.506 (5)          & 0.484 (7)          & 0.528 (2)          & 0.515 (4)          & \textbf{0.525 (3)} & 0.42 (12)     & 0.458 (11)         & 0.48 (9)           & 0.478 (10)     & 0.484 (7)          & 0.487 (6)          & \textbf{0.53 (1)}  \\
Shuttle (arff)          & 0.263 (5)          & 0.358 (3)          & 0.08 (12)          & 0.13 (10)          & 0.36 (2)           & 0.132 (9)     & \textbf{0.395 (1)} & 0.184 (7)          & 0.304 (4)      & 0.232 (6)          & 0.171 (8)          & 0.085 (11)         \\
SpamBase (arff)         & 0.38 (10)          & 0.414 (8)          & 0.525 (2)          & 0.492 (3)          & 0.422 (6)          & 0.375 (11)    & 0.349 (12)         & 0.45 (4)           & 0.412 (9)      & 0.42 (7)           & 0.431 (5)          & \textbf{0.567 (1)} \\
Stamps (arff)           & 0.247 (10)         & 0.241 (11)         & 0.398 (5)          & 0.394 (6)          & 0.341 (9)          & 0.404 (4)     & 0.229 (12)         & 0.425 (2)          & 0.346 (8)      & 0.411 (3)          & 0.392 (7)          & \textbf{0.464 (1)} \\
Waveform (arff)         & 0.066 (8)          & 0.129 (2)          & 0.055 (10)         & 0.056 (9)          & \textbf{0.134 (1)} & 0.052 (12)    & 0.108 (4)          & 0.114 (3)          & 0.07 (6)       & 0.054 (11)         & 0.07 (6)           & \textbf{0.079 (5)} \\
WBC (arff)              & 0.542 (11)         & 0.556 (9)          & 0.694 (6)          & \textbf{0.863 (1)} & 0.581 (8)          & 0.724 (4)     & 0.33 (12)          & 0.543 (10)         & 0.702 (5)      & 0.638 (7)          & 0.728 (3)          & 0.838 (2)          \\
WDBC (arff)             & 0.43 (12)          & 0.667 (9)          & 0.795 (2)          & 0.718 (6)          & 0.654 (10)         & 0.794 (3)     & 0.704 (7)          & 0.773 (4)          & 0.612 (11)     & 0.688 (8)          & 0.729 (5)          & \textbf{0.841 (1)} \\
WPBC (arff)             & 0.21 (12)          & 0.224 (8)          & 0.23 (6)           & 0.231 (5)          & 0.233 (4)          & 0.223 (9)     & 0.225 (7)          & \textbf{0.248 (1)} & 0.223 (9)      & 0.223 (9)          & 0.245 (2)          & \textbf{0.244 (3)} \\
\midrule
AVG                     & 0.351 (12)         & 0.462 (8)          & 0.509 (3)          & 0.538 (2)          & 0.447 (9)          & 0.436 (10)    & 0.378 (11)         & 0.466 (7)          & 0.49 (6)       & 0.503 (4)          & 0.503 (4)          & \textbf{0.565 (1)}
 \\
\bottomrule
\end{tabular}
}
	\label{table:od comparison map} % is used to refer this table in the text
\end{table*}
\begin{table}[!htb]
\caption{\method's runtime (in seconds) on a moderate laptop under different sample size ($n$) and dimensions ($d$). It scales well to handle high-dimensional, large datasets.}
% \vspace{-0.1in}
\centering

\begin{tabular}{l|llll}
    \toprule
     & d=10 & d=100 & d=1,000 & d=10,000 \\
    \midrule
    n=1,000 & 0.068 & 0.173 & 1.163 & 11.460 \\
    n=10,000 & 0.171 & 0.468 & 5.244 & 55.190 \\
    n=100,000 & 0.640 & 7.185 & 70.541 & 567.105 \\
    n=1,000,000 & 11.403 & 130.974 & 694.405 & 5376.593\\
    \bottomrule
\end{tabular}
\label{table:copod speed}
\end{table}
\begin{table*}[!htb]
\setlength{\tabcolsep}{4pt}
\centering
	\caption{ Run time (in seconds) of detectors (average of 10 independent trials, the fastest is highlighted in bold); rank is shown in parenthesis (lower is better). \method is one of the fastest algorithms.}
\footnotesize
\scalebox{1}{
\scriptsize
\begin{tabular}{l|lllllllllll|l}
\toprule
\textbf{Data}           & \textbf{ABOD}      & \textbf{CBLOF}     & \textbf{HBOS}      & \textbf{IForest}   & \textbf{KNN}       & \textbf{LODA} & \textbf{LOF}       & \textbf{LSCP} & \textbf{OCSVM} & \textbf{PCA}       & \textbf{SUOD}      & \textbf{ECOD}      \\
\midrule
Arrhythmia (mat)        & 0.236 (7)          & 0.258 (8)          & \textbf{0.201 (6)} & \textbf{0.306 (10)} & 0.077 (5)          & 0.05 (2)           & 0.066 (4)          & 0.984 (11)          & \textbf{0.042 (1)} & 0.058 (3)          & 2.004 (12)          & 0.266 (9)          \\
Breastw (mat)           & 0.17 (9)           & 0.076 (8)          & 0.004 (2)          & 0.384 (10)          & 0.038 (7)          & 0.037 (6)          & 0.007 (3)          & 0.524 (11)          & 0.01 (4)           & \textbf{0.002 (1)} & 1.431 (12)          & \textbf{0.024 (5)} \\
Cardio (mat)            & 0.546 (10)         & 0.175 (8)          & 0.011 (2)          & 0.426 (9)           & 0.158 (7)          & 0.06 (4)           & 0.116 (6)          & 1.512 (11)          & 0.086 (5)          & \textbf{0.005 (1)} & 1.811 (12)          & 0.053 (3)          \\
Ionosphere (mat)        & 0.083 (9)          & 0.053 (8)          & 0.012 (4)          & 0.271 (10)          & \textbf{0.015 (5)} & 0.025 (6)          & 0.006 (3)          & 0.45 (11)           & 0.004 (2)          & \textbf{0.003 (1)} & 1.458 (12)          & 0.047 (7)          \\
Lympho (mat)            & 0.036 (8)          & 0.051 (9)          & \textbf{0.009 (5)} & \textbf{0.253 (10)} & 0.008 (4)          & 0.023 (6)          & 0.003 (3)          & 0.32 (11)           & \textbf{0.001 (1)} & 0.002 (2)          & 1.399 (12)          & 0.036 (8)          \\
Mammography (mat)       & 1.701 (10)         & 0.268 (6)          & \textbf{0.006 (1)} & 0.689 (8)           & 0.473 (7)          & 0.091 (4)          & 0.255 (5)          & 4.6 (12)            & 1.364 (9)          & 0.007 (2)          & 2.517 (11)          & \textbf{0.044 (3)} \\
Optdigits (mat)         & 2.154 (10)         & 0.433 (5)          & \textbf{0.032 (1)} & 0.725 (6)           & 1.457 (9)          & 0.063 (3)          & 1.371 (8)          & 14.566 (12)         & 1.169 (7)          & 0.046 (2)          & 7.114 (11)          & 0.229 (4)          \\
Pima (mat)              & 0.176 (9)          & 0.072 (8)          & \textbf{0.001 (1)} & 0.267 (10)          & \textbf{0.03 (7)}  & 0.023 (6)          & 0.009 (4)          & 0.577 (11)          & 0.007 (3)          & 0.003 (2)          & 1.452 (12)          & 0.023 (6)          \\
Satellite (mat)         & 1.602 (10)         & \textbf{0.399 (5)} & \textbf{0.018 (1)} & 0.669 (6)           & 0.867 (8)          & 0.071 (3)          & 0.835 (7)          & 8.549 (12)          & 1.075 (9)          & 0.024 (2)          & 5.173 (11)          & 0.158 (4)          \\
Satimage-2 (mat)        & 1.43 (10)          & \textbf{0.354 (5)} & 0.017 (2)          & 0.546 (6)           & 0.667 (8)          & 0.063 (3)          & 0.64 (7)           & 7.699 (12)          & 0.91 (9)           & 0.017 (2)          & 4.742 (11)          & 0.146 (4)          \\
Shuttle (mat)           & 2.01 (10)          & 0.175 (4)          & 0.006 (2)          & \textbf{0.634 (6)}  & 0.788 (8)          & 0.076 (3)          & 0.728 (7)          & 7.973 (12)          & 1.33 (9)           & \textbf{0.003 (1)} & 3.566 (11)          & \textbf{0.189 (5)} \\
Speech (mat)            & \textbf{7.85 (10)} & 2.036 (5)          & 0.167 (2)          & 3.589 (6)           & 7.842 (9)          & \textbf{0.154 (1)} & 7.274 (8)          & 49.828 (12)         & 4.272 (7)          & 1.052 (3)          & 14.31 (11)         & 1.148 (4)          \\
Wbc (mat)               & 0.08 (9)           & 0.06 (8)           & 0.008 (4)          & 0.239 (10)          & 0.015 (5)          & 0.021 (6)          & 0.007 (3)          & 0.47 (11)           & 0.004 (2)          & \textbf{0.002 (1)} & 1.495 (12)          & \textbf{0.055 (7)} \\
Wine (mat)              & 0.026 (8)          & 0.046 (9)          & 0.005 (5)          & 0.243 (10)          & 0.005 (5)          & 0.02 (7)           & 0.002 (3)          & 0.313 (11)          & \textbf{0.001 (1)} & 0.002 (3)          & 1.421 (12)          & \textbf{0.019 (6)} \\
Arrhythmia (arff)       & 0.226 (8)          & 0.23 (9)           & 0.169 (6)          & 0.249 (10)          & 0.065 (4)          & 0.043 (2)          & 0.057 (3)          & 0.947 (11)          & \textbf{0.039 (1)} & 0.095 (5)          & \textbf{2.025 (12)} & \textbf{0.226 (8)} \\
Cardiotocography (arff) & 0.368 (10)         & 0.124 (7)          & 0.007 (2)          & 0.286 (9)           & 0.138 (8)          & 0.049 (4)          & 0.099 (6)          & 1.769 (11)          & 0.084 (5)          & \textbf{0.007 (2)} & 1.876 (12)          & 0.037 (3)          \\
HeartDisease (arff)     & 0.048 (9)          & 0.044 (8)          & \textbf{0.005 (4)} & 0.21 (10)           & 0.008 (5)          & 0.018 (7)          & 0.004 (3)          & 0.366 (11)          & \textbf{0.001 (1)} & 0.002 (2)          & 1.38 (12)           & \textbf{0.016 (6)} \\
Hepatitis (arff)        & 0.013 (6)          & 0.028 (9)          & 0.004 (5)          & 0.187 (10)          & 0.003 (4)          & 0.016 (7)          & 0.002 (3)          & 0.292 (11)          & 0.001 (2)          & 0.001 (2)          & 1.378 (12)          & \textbf{0.021 (8)} \\
InternetAds (arff)      & 5.648 (9)          & 1.36 (3)           & \textbf{0.493 (2)} & 5.605 (8)           & 5.252 (6)          & \textbf{0.227 (1)} & 6.11 (10)          & 58.435 (12)         & 4.159 (5)          & 5.496 (7)          & 21.974 (11)         & 1.855 (4)          \\
Ionosphere (arff)       & \textbf{0.075 (9)} & 0.049 (8)          & 0.012 (4)          & 0.236 (10)          & \textbf{0.014 (5)} & 0.02 (6)           & 0.007 (3)          & 0.447 (11)          & 0.004 (2)          & 0.004 (2)          & 1.498 (12)          & 0.036 (7)          \\
KDDCup99 (arff)         & 136.469 (9)        & \textbf{2.1 (5)}   & \textbf{0.135 (1)} & \textbf{7.714 (6)}  & 133.6 (8)        & 0.409 (3)          & 141.9 (10)       & 839.8 (12)        & 175.1 (11)       & \textbf{0.301 (2)} & 68.726 (7)          & \textbf{0.965 (4)} \\
Lymphography (arff)     & 0.039 (8)          & 0.053 (9)          & 0.007 (5)          & \textbf{0.255 (10)} & 0.007 (5)          & 0.023 (7)          & 0.003 (3)          & 0.31 (11)           & \textbf{0.001 (1)} & 0.002 (2)          & 1.503 (12)          & \textbf{0.02 (6)}  \\
Pima (arff)             & 0.201 (9)          & 0.087 (8)          & 0.004 (2)          & 0.319 (10)          & \textbf{0.034 (6)} & 0.029 (5)          & 0.014 (4)          & 0.573 (11)          & 0.013 (3)          & \textbf{0.002 (1)} & 1.43 (12)           & \textbf{0.043 (7)} \\
Shuttle (arff)          & 0.248 (8)          & 0.079 (7)          & 0.004 (2)          & 0.366 (9)           & 0.05 (6)           & 0.033 (5)          & \textbf{0.023 (4)} & 0.712 (10)          & 0.018 (3)          & \textbf{0.003 (1)} & 1.481 (12)          & 0.912 (11)         \\
SpamBase (arff)         & 1.329 (10)         & 0.303 (5)          & \textbf{0.017 (1)} & 0.514 (6)           & 0.904 (9)          & 0.061 (3)          & 0.888 (8)          & 8.349 (12)          & 0.666 (7)          & 0.032 (2)          & 3.836 (11)          & \textbf{0.109 (4)} \\
Stamps (arff)           & 0.071 (9)          & 0.055 (8)          & 0.004 (4)          & 0.233 (10)          & 0.013 (5)          & 0.018 (7)          & 0.003 (3)          & 0.384 (11)          & 0.002 (2)          & 0.002 (2)          & 1.382 (12)          & \textbf{0.014 (6)} \\
Waveform (arff)         & 1.018 (10)         & 0.244 (5)          & 0.01 (2)           & 0.478 (8)           & \textbf{0.521 (9)} & 0.072 (4)          & 0.429 (7)          & 3.479 (12)          & 0.261 (6)          & \textbf{0.008 (1)} & 2.357 (11)          & \textbf{0.047 (3)} \\
WBC (arff)              & 0.048 (8)          & 0.05 (9)           & 0.004 (4)          & \textbf{0.243 (10)} & 0.009 (5)          & 0.021 (7)          & 0.003 (3)          & 0.335 (11)          & 0.001 (2)          & 0.001 (2)          & 1.367 (12)          & 0.013 (6)          \\
WDBC (arff)             & 0.092 (9)          & 0.062 (8)          & 0.009 (4)          & 0.245 (10)          & 0.016 (5)          & 0.022 (6)          & 0.007 (3)          & 0.466 (11)          & 0.005 (2)          & \textbf{0.004 (1)} & 1.5 (12)            & \textbf{0.032 (7)} \\
WPBC (arff)             & 0.046 (7)          & 0.052 (8)          & 0.01 (5)           & 0.237 (10)          & 0.008 (4)          & 0.021 (6)          & 0.004 (3)          & \textbf{0.353 (11)} & \textbf{0.002 (1)} & 0.003 (2)          & 1.383 (12)          & \textbf{0.062 (9)} \\
\midrule
AVG                     & 5.468 (9)          & 0.312 (5)          & \textbf{0.046 (1)} & 0.887 (6)           & 5.105 (7)          & 0.062 (2)          & 5.362 (8)          & 33.847 (12)         & 6.355 (11)         & 0.24 (4)           & 5.5 (10)            & \textbf{0.228 (3)}\\
\bottomrule
\end{tabular}}
	\label{table:od comparison time} % is used to refer this table in the text
\end{table*}

\begin{figure}[!htb]
\centering
\includegraphics[width=0.45\textwidth]{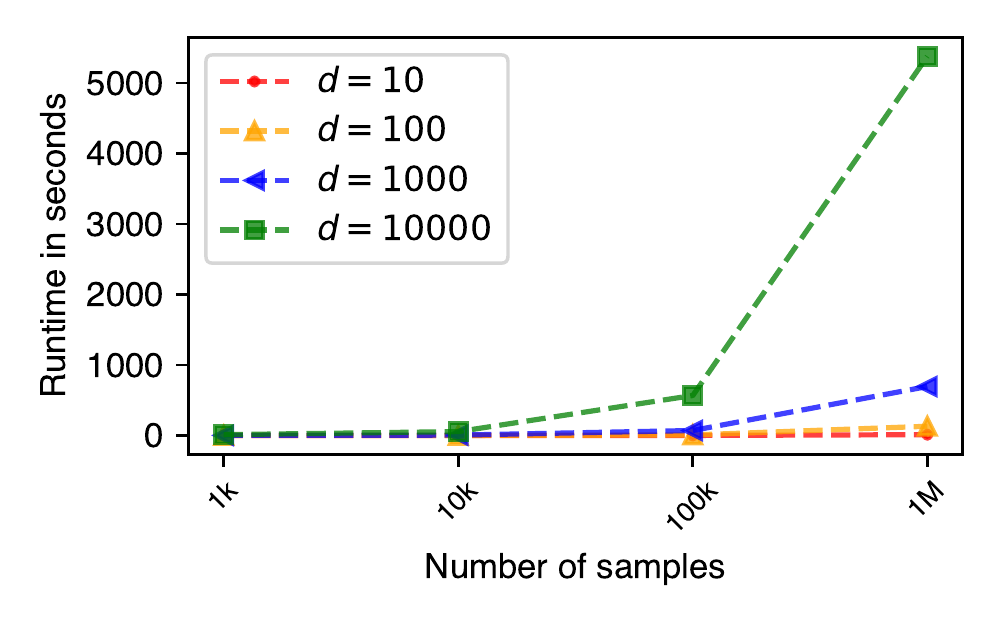}
\caption{Runtime of \method on the synthetic datasets with a varying number of samples and dimensions. It is efficient and scalable to handle a million samples with 10,000 dimensions in less than 2 hours.
}
\label{fig:scalability}
\end{figure}

\subsection{Runtime Efficiency and Scalability}
\label{subsec:scalability}

\textbf{\method is an efficient algorithm with fast computation}. Table \ref{table:od comparison time} shows that of the 12 algorithms tested, \method ranks third in runtime---an average of 0.228 seconds to process each dataset. The fast two algorithms are HBOS and LODA, which are known to be efficient by building uni-dimensional histograms. Intuitively, \method's  efficiency is attributed to the feature independence assumption.

\textbf{\method is also a scalable algorithm that suits high dimensional settings}. Unlike proximity-based models that require extensive distance calculation or density estimation, \method incurs very little computational overhead. We carry out the experiment by generating synthetic datasets of dimensions 10, 100, 1,000, and 10,000 and the number of observations are 1,000, 10,000, 100,000, and 1,000,000. All datasets are randomly generated and used strictly for evaluating \method's computation time. As such, we ignore the performance metrics and only keep performance time. Fig. \ref{fig:scalability} and Table \ref{table:copod speed} illustrate  \method's performance and its scalability across varying the number of samples ($n$) and dimensions ($d$). Even on the moderately equipped personal computer, \method can easily handle datasets with 10,000 dimensions and 1,000,000 data points in 2 hours. The result is consistent with our complexity analysis in Section \ref{subsubsec:complexity}---\method has $\mathcal{O}(nd)$ time complexity that scales linearly in both the number of samples and dimensions.

Additionally, \method requires no re-training to fit new data points with relatively large data samples if no data shift is assumed, and thus is highly desirable for applications where real-time predictions are needed. Within PyOD, we also provide the distributed implementation with even more efficiency and scalability.

\section{Conclusion and Future Works}
\label{sec:conclusion}
In this paper, we present a novel outlier detection method called \method that uses empirical cumulative distribution functions in measuring the outlyingness of data points. Specifically, it computes left- and right-tail univariate ECDFs per dimension. For every data point, \method uses the univariate ECDFs to estimate tail probabilities for the data point and aggregates these tail probabilities to come up with a final outlier score. The intuition of \method somewhat resembles how p-value works: for a specific data point, we are looking at how low its tail probability is.
The proposed \method is parameter-free without the need for hyperparameter tuning. Our extensive evaluation on 30 benchmark datasets shows that \method outperforms SOTA baselines, while being a fast and scalable outlier detection algorithm. We provide an easy-to-use Python implementation of \method for both single- and multi-thread support. We leave accounting for how features are related for future work. For instance, we can learn vine copulas \cite{joe2011dependence} or probabilistic graphical models \cite{hastie2019statistical} to identify blocks of features that are highly correlated, and only for these blocks (rather than across the entire dataset), we could compute a joint ECDF. We suspect that these strategies would require some amount of hyperparameter tuning though. Separately, our approach is not designed to handle multimodal distributions for which an outlier could be in neither left nor right tails. Figuring out a way to extend \method to such settings while still being fast and scalable is also a possible direction worth exploring.

% Can use something like this to put references on a page
% by themselves when using endfloat and the captionsoff option.
\ifCLASSOPTIONcaptionsoff
  \newpage
\fi

% trigger a \newpage just before the given reference
% number - used to balance the columns on the last page
% adjust value as needed - may need to be readjusted if
% the document is modified later
%\IEEEtriggeratref{8}
% The "triggered" command can be changed if desired:
%\IEEEtriggercmd{\enlargethispage{-5in}}

% references section

% can use a bibliography generated by BibTeX as a .bbl file
% BibTeX documentation can be easily obtained at:
% http://mirror.ctan.org/biblio/bibtex/contrib/doc/
% The IEEEtran BibTeX style support page is at:
% http://www.michaelshell.org/tex/ieeetran/bibtex/
%\bibliographystyle{IEEEtran}
% argument is your BibTeX string definitions and bibliography database(s)
%\bibliography{IEEEabrv,../bib/paper}
%
% <OR> manually copy in the resultant .bbl file
% set second argument of \begin to the number of references
% (used to reserve space for the reference number labels box)
% \begin{thebibliography}{1}

% \bibitem{IEEEhowto:kopka}
% H.~Kopka and P.~W. Daly, \emph{A Guide to \LaTeX}, 3rd~ed.\hskip 1em plus
%   0.5em minus 0.4em\relax Harlow, England: Addison-Wesley, 1999.

% \end{thebibliography}
\bibliographystyle{IEEEtran}
\bibliography{ecod}
% \clearpage
% \newpage

% biography section
% 
% If you have an EPS/PDF photo (graphicx package needed) extra braces are
% needed around the contents of the optional argument to biography to prevent
% the LaTeX parser from getting confused when it sees the complicated
% \includegraphics command within an optional argument. (You could create
% your own custom macro containing the \includegraphics command to make things
% simpler here.)
%\begin{IEEEbiography}[{\includegraphics[width=1in,height=1.25in,clip,keepaspectratio]{mshell}}]{Michael Shell}
% or if you just want to reserve a space for a photo:

% \begin{IEEEbiography}[{\includegraphics[width=1in,height=1.25in,clip,keepaspectratio]{mshell}}]{Michael Shell}
% \end{IEEEbiography}

\begin{IEEEbiography}[{\includegraphics[width=1in,height=1.25in,clip,keepaspectratio]{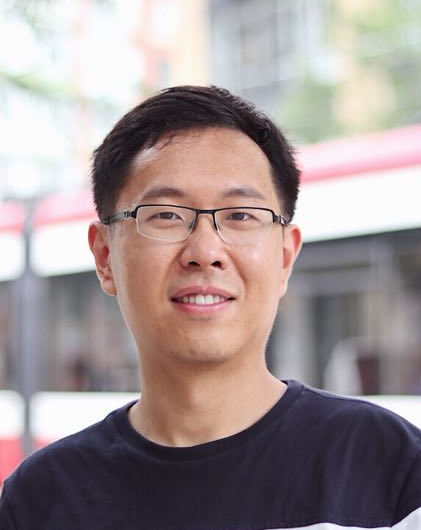}}]{Zheng Li}
is the founder of Arima, a Toronto based data mining startup and an adjunct lecturer at Northeastern University Toronto.
\end{IEEEbiography}

\begin{IEEEbiography}[{\includegraphics[width=1in,height=1.25in,clip,keepaspectratio]{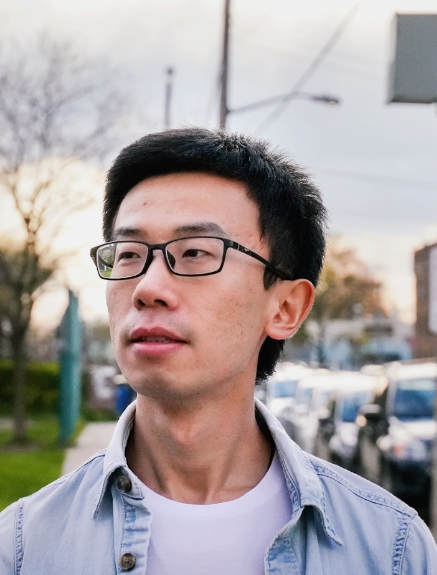}}]{Yue Zhao}
is a Ph.D. student at Carnegie Mellon University (CMU), focusing on anomaly detection (a.k.a outlier detection) algorithms, systems, and its applications in security, healthcare, and Finance, with more than 7-year professional experience and 20+ papers (in JMLR, TKDE, NeurIPS, etc.). He is a recipient of the 2022 Norton Labs Graduate Fellowship (formerly known as the Symantec Research Labs Fellowship).
\end{IEEEbiography}

\begin{IEEEbiography}[{\includegraphics[width=1in,height=1.25in,clip,keepaspectratio]{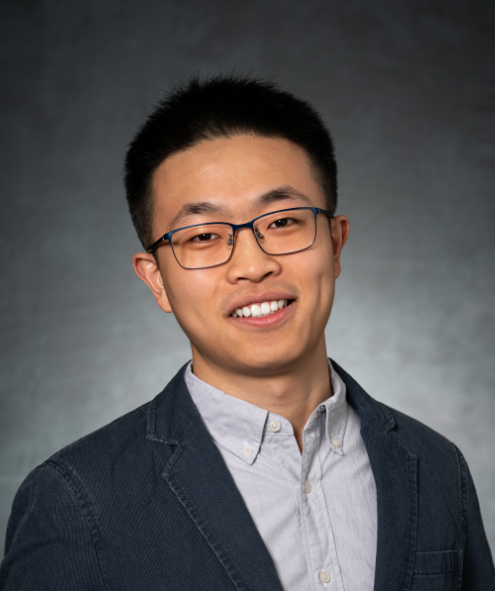}}]{Xiyang Hu}
is a Ph.D. student at Carnegie Mellon University. He got his M.Sc. in Statistical Science from Duke University, and B.Arch. in Architecture with a minor in Computer Science from Tsinghua University. His research interests include machine learning, deep learning, data mining, and social impacts of AI.
\end{IEEEbiography}

\begin{IEEEbiography}
[{\includegraphics[width=1in,height=1.25in,clip,keepaspectratio]{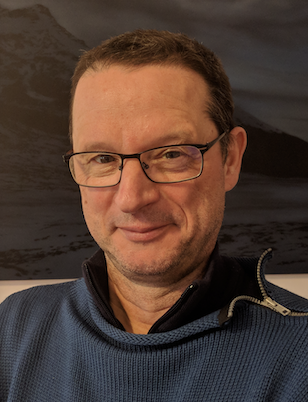}}]
{Nicola Botta}
has worked on numerical methods for conservation laws,
agent-based modelling and verified decision making. He is a senior
scientist at the Potsdam Institute for Climate Impact Research, Potsdam,
Germany and adjunct professor in the Functional Programming division at
the Computer Science and Engineering Department, Chalmers University of
Technology, Sweden.
\end{IEEEbiography}

\begin{IEEEbiography}[{\includegraphics[width=1in,height=1.25in,clip,keepaspectratio]{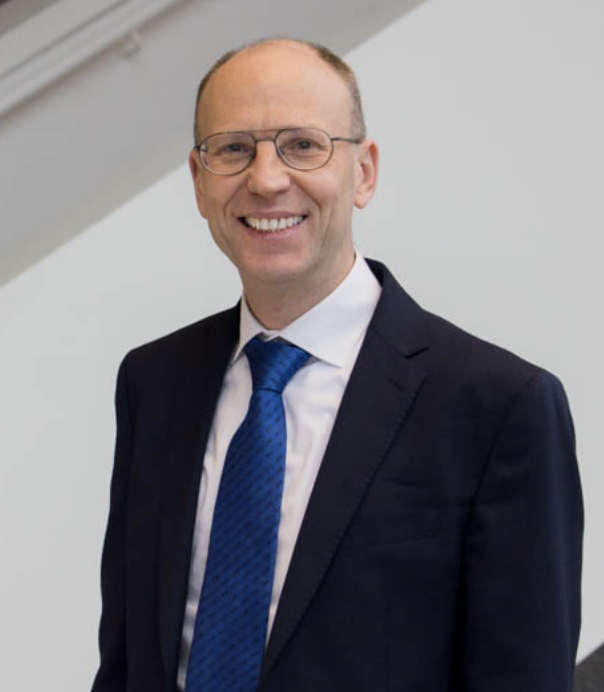}}]{Cezar Ionescu}
is Professor of Computer Science at Technische Hochschule Deggendorf, Deggendorf, Germany.
\end{IEEEbiography}

\begin{IEEEbiography}[{\includegraphics[width=1in,height=1.25in,clip,keepaspectratio]{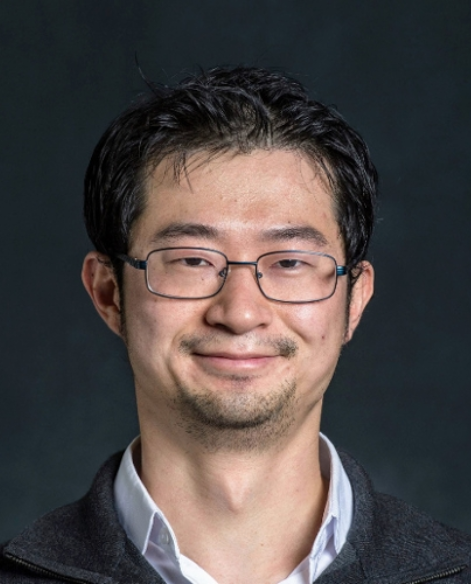}}]{George H.~Chen}
is an assistant professor of information systems at Carnegie Mellon University. He primarily works on building trustworthy machine learning models for time-to-event prediction (survival analysis) and for time series analysis. He often uses nonparametric prediction models that work well under very few assumptions on the data. His main application area is in healthcare.
\end{IEEEbiography}

% % if you will not have a photo at all:
% \begin{IEEEbiographynophoto}{Yue Zhao}
% Biography text here.
% \end{IEEEbiographynophoto}

% insert where needed to balance the two columns on the last page with
% biographies
% \newpage

% \begin{IEEEbiographynophoto}{Jane Doe}
% Biography text here.
% \end{IEEEbiographynophoto}

% You can push biographies down or up by placing
% a \vfill before or after them. The appropriate
% use of \vfill depends on what kind of text is
% on the last page and whether or not the columns
% are being equalized.

%\vfill

% Can be used to pull up biographies so that the bottom of the last one
% is flush with the other column.
\enlargethispage{-6.38in}

% that's all folks
\end{document}